\documentclass[10pt]{article} %
\usepackage[preprint]{tmlr}

\usepackage{amsmath,amsfonts,bm}

\def\eqref#1{equation~\ref{#1}}

\def\1{\bm{1}}

\DeclareMathAlphabet{\mathsfit}{\encodingdefault}{\sfdefault}{m}{sl}
\SetMathAlphabet{\mathsfit}{bold}{\encodingdefault}{\sfdefault}{bx}{n}

\usepackage[utf8]{inputenc} %
\usepackage[T1]{fontenc}    %
\usepackage[bottom]{footmisc} %
\usepackage{hyperref}       %
\hypersetup{hidelinks}
\usepackage{url}            %
\usepackage{booktabs}       %
\usepackage{amsfonts}       %
\usepackage{nicefrac}       %
\usepackage{microtype}      %

\usepackage{graphicx}
\usepackage{amssymb}
\usepackage{amsmath}
\usepackage{enumitem}
\usepackage{subcaption}
\usepackage{xcolor}
\usepackage{colortbl}
\usepackage{rotating}
\usepackage{pifont}
\usepackage{bbding}
\definecolor{pairwisesignificant}{RGB}{250,218,221}

\newcommand{\new}{\textcolor{orange}{{\(\star\)}}}

\usepackage{soul}
\sethlcolor{yellow}

\newcommand{\newmodel}{$\chi$ViT}
\title{GeoCrossBench: Cross-Band Generalization \\ for Remote Sensing}

\author{\name Hakob Tamazyan, Ani Vanyan, Alvard Barseghyan, Anna Khosrovyan
      \email \{hakob,ani,alla,anna\}@yerevann.com \\
      \addr YerevaNN and Yerevan State University, Yerevan, Armenia
      \AND
      \name Evan Shelhamer \email shelhamer@cs.ubc.ca \\
      \addr University of British Columbia and Vector Institute, Canada
      \AND
      \name Hrant Khachatrian \email hrant@yerevann.com \\
      \addr YerevaNN and Yerevan State University, Yerevan, Armenia}

\def\month{09}
\def\year{2026}
\def\openreview{\url{https://openreview.net/forum?id=vmcNMXWzpQ}}
\begin{document}

\maketitle

\begin{abstract}
The data for remote sensing is constantly acquired, and new data comes from a growing number and diversity of satellites, while the vast majority of labeled data comes from older satellites.
As remote-sensing foundation models for Earth observation scale up, the cost of \mbox{(re-)}training to support new satellites grows too, so cross-band generalization across sensors and satellites is increasingly important.
We introduce GeoCrossBench, an extension of the popular GeoBench benchmark with a new evaluation protocol for cross-band generalization across sensors and satellites: it tests standard in-distribution performance with the same bands for train and test, generalization to inputs with no intersection between train and test; and generalization to test inputs containing a superset of the training bands.
We develop \newmodel{}, a self-supervised extension of the band-agnostic ChannelViT, as a supporting baseline for cross-band generalization.
We evaluate a representative set of remote-sensing-specific and general-purpose vision models, characterize current performance, and identify directions for improvement through 11,900 H100 GPU-hours of experiments.
When averaging dataset-specific metric scores, DOFA leads the in-distribution setting (61.30), frozen Panopticon leads the no-overlap setting (22.75), and ImageNet-pretrained ViT-B leads both the superset setting (56.19) and the overall average across settings (45.27). While top rankings in each setting are close, we clearly see that all models suffer significant performance losses when evaluated on unseen bands.
We will publicly release the code and datasets to support the development of more future-proof remote sensing models with stronger cross-band generalization.
\end{abstract}

\section{Introduction: Generalization across Remote Sensing Data}

The growth of remote sensing data and satellite imagery in particular \citep{earthengine, zhu2017deep, ma2019deep} has led to the development of sophisticated deep learning models capable of analyzing complex geospatial patterns and dynamics.
Among these, pre-trained foundation models have emerged as a popular paradigm for learning generalizable representations from vast and diverse remote sensing (RS) datasets \citep{dofa, croma, terramind, satmae, msGFM, galileo, terrafm, prithvi, softcon}.
RS data is not only spatiotemporal, but inherently multimodal, capturing diverse spectral \emph{bands} including multispectral, hyperspectral, and synthetic aperture radar (SAR) \citep{sen1, sen2, landsat8, enmap, missioncritical}.
These foundation models promise ease-of-use and transfer across RS data.
We examine how well they transfer across modalities, in the forms of bands/channels, sensors, and satellites.

While current models can transfer well across space and time, when train and test bands match, we find that their \textbf{cross-band generalization remains limited}, when applied to bands and sensors unseen during training and fine-tuning.
Retraining on the new data may be effective, but is highly costly, and all the more costly as models continue to grow.
This issue is practically important, because this type of generalization determines how well a model applies across different spectra and modalities, such as from optical RGB to optical multi-spectral imagery or active radar (SAR) imagery.
This different data may be needed, to improve the accuracy of an analysis by incorporating a complementary satellite, or to switch an analysis to a new satellite when the old satellite is decomissioned.

This is a critical gap: real-world applications can require models to summarize data from various sensors, to adapt to new bands, or to do a new task that needs bands complementary to the training bands.
Robust generalization across bands is crucial for creating more versatile and practical remote sensing models, because large-scale training and fine-tuning is not accessible for all researchers and practitioners.

We introduce \textbf{GeoCrossBench} to assess the gap of cross-band generalization in remote sensing with \emph{three} complementary evaluation protocols: (1) \emph{in-distribution} -- train and test on the same bands, (2) \emph{no overlap bands generalization} -- train on optical bands and test on not overlapping bands, and (3) \emph{superset bands generalization} -- test-time inputs provide strictly more bands than used in training.

\begin{figure}[t]
 \centering
 \includegraphics[clip, trim=2.5cm 20.5cm 6.8cm 2.5cm, width=0.96\textwidth]
 {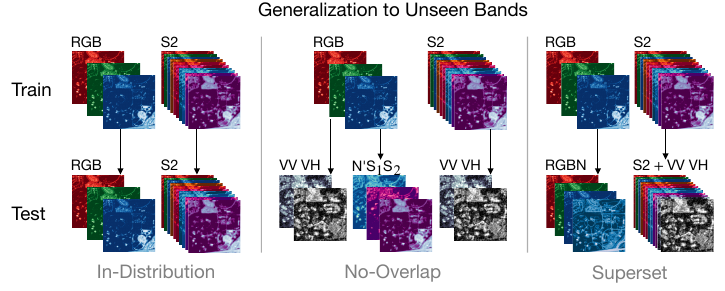}
 \caption{%
 The GeoCrossBench evaluation framework.
 (1) \emph{In-Distribution}: fine-tune on RGB and evaluate on RGB; fine-tune on full S2 and evaluate on S2. (2) \emph{No-Overlap}: evaluate transfer from RGB$\to$S1 (VV, VH), RGB$\to$N'S\textsubscript{1}S\textsubscript{2} (B8A, B11, B12) and S2$\to$S1. (3) \emph{Superset}: RGB$\to$RGBN (RGB+NIR) and S2$\to$S2+S1 (optical+SAR fusion).
 }
 \label{fig:geocrossbench}
\end{figure}

GeoCrossBench covers three canonical remote sensing tasks: scene classification, semantic segmentation, and change detection, covering both Sentinel-2 (S2) optical/multispectral data and Sentinel-1 (S1) SAR data.
Specifically, we build GeoCrossBench from the GeoBench datasets~\citep{geobench} and enrich them with additional public datasets that widen the range of resolutions and geographic contexts.
Moreover, for the datasets missing SAR bands we fuse the Sentinel-2 multispectral bands with co-registered Sentinel-1 SAR bands (VV/VH dual-polarization).
This fusion expands the spectral range of the datasets to allow for more rigorous cross-band evaluation.
The core idea of GeoCrossBench is to train models on a common band configuration (e.g., RGB, S2) and then evaluate on a variety of unseen bands from both optical and SAR modalities, as illustrated in Figure~\ref{fig:geocrossbench}.
To provide a comprehensive analysis that also considers practical computational constraints, we evaluate generalization using two adaptation regimes: full fine-tuning and fine-tuning with frozen backbone.

GeoCrossBench and the systematic evaluation it enables are our primary contribution; we evaluate 8 remote-sensing-specific and 5 general-purpose vision models, requiring 11,900 H100 GPU-hours.
As a supporting contribution, building on ChannelViT \citep{channelvit}, an extension of the Vision Transformer (ViT) \citep{ViT} for channel-wise modeling, we develop a new baseline for band-wise modeling in RS. We call this model \textbf{\newmodel{}} (ChiViT), short for \textbf{Ch}annel-based \textbf{i}BOT pre-trained \textbf{ViT}, and pretrain it using the iBOT \citep{ibot} paradigm on our own large-scale, multi-modal dataset.

Experiments with our benchmark reveal insights into current performance and potential directions of improvement.
We find that many foundation models struggle with cross-band generalization.
The aggregate results have different leaders by setting: DOFA in-distribution, frozen Panopticon in the no-overlap setting, and ViT-B for the superset setting and overall.
Finally, we find that \newmodel{} is a strong supporting baseline for no-overlap transfer, while Panopticon is a close competitor and general-purpose models lead other settings. These findings underscore the need for a rigorous and standardized benchmark like GeoCrossBench.

We will publicly release the GeoCrossBench code and datasets.
This release can help measure progress, identify weaknesses in current approaches, and ultimately drive the development of more robust, versatile, and reliable foundation models for comprehensive Earth observation.

\section{GeoCrossBench Benchmark: Dataset and Evaluation Protocol}
GeoCrossBench is designed to thoroughly evaluate the ability of remote sensing foundation models to generalize knowledge learned from one set of spectral bands (specifically RGB or S2) to other spectral band combinations that either match, do not overlap with, or strictly supersede the training bands.

\subsection{Motivation and Design Principles}

The primary motivation behind GeoCrossBench is the observation that many foundation models achieve high performance when training and testing data come from the same spectral distribution, yet their performance often degrades when processing imagery with different spectral characteristics. This limitation hinders their practical utility. This generalization challenge is not a theoretical concern; it poses a significant barrier to deploying models in a constantly evolving satellite ecosystem where large-scale labeled datasets are often unavailable for newer commercial satellites. A practitioner might need to transfer a model trained on public Sentinel-2 data to newer platforms like Planet SuperDove or Satlantis GARAI, which share some spectral bands with Sentinel-2 but also introduce new ones (e.g., Green I, Yellow). More extreme generalization is required when transferring between entirely different sensor types. For example, adapting an optical model from Sentinel-2 for use with SatVu's HotSat, which captures purely thermal data, requires generalization to a non-overlapping spectral range. The same challenge arises when transferring from optical to SAR imagery. GeoCrossBench directly evaluates one cross-satellite transition: transfer from Sentinel-2 optical imagery to Sentinel-1 SAR imagery. The remaining within-Sentinel-2 transformations serve as proxies for transitions to satellites such as Planet SuperDove, Satlantis GARAI, or SatVu HotSat that provide different or additional spectral bands.

GeoCrossBench is built on the following principles:
\begin{itemize}[leftmargin=*]
    \item \textbf{Focus on Generalization:} The main goal is to evaluate how well models adapt from a seen spectral inputs to unseen ones.
    \item \textbf{RS Specific Tasks:} Evaluation is based on tasks central to remote sensing: scene classification, semantic segmentation, and change detection.
    \item \textbf{Diverse Spectral Modalities:} The benchmark incorporates both multi-band optical data and dual-polarization SAR data to test generalization across fundamentally different sensing mechanisms.
\end{itemize}

\subsection{Datasets}

GeoCrossBench extends the original GeoBench benchmark by fusing them with corresponding Sentinel-1 SAR data and also incorporates completely new datasets relevant to cross-band generalization, such as x-sen1floods11, x-oscd, x-harvey-flood and x-harvey-building. All datasets in GeoCrossBench utilize Sentinel-2 10-band optical data (B2, B3, B4, B5, B6, B7, B8, B8A, B11, B12 -- bands with $\leq$20m resolution) and Sentinel-1 dual-polarization SAR data (VV, VH), resulting in a 12-band input for each sample. An overview of the datasets is provided in Table~\ref{tab:geocrossbench_datasets}.
The difference between the x-harvey dataset and the original \citep{harvey} lies in the split we provide, which bypasses the geographical distribution shift. Additionally, we use the original dataset to construct a change detection task by pairing pre- and post-flood images, along with the corresponding flood segmentation masks.
x-sen1floods11 is a subset of the original Sen1Floods11 dataset \citep{sen1floods11}, created by removing the weakly labeled portion.

\begin{table}[h!]
\centering
\caption{Overview of the datasets included in GeoCrossBench. The ones marked with \new\ are not part of the original GeoBench.}
\label{tab:geocrossbench_datasets}
\resizebox{\textwidth}{!}{
\begin{tabular}{@{}llclrrr@{}}
\toprule
\textbf{Dataset Name} & \textbf{Image Size} & \textbf{\#Classes} & \textbf{Sensors/Bands} & \textbf{Train} & \textbf{Val} & \textbf{Test} \\
\midrule[\heavyrulewidth]
\multicolumn{7}{c}{\textit{Classification}} \\
\midrule
x-bigearthnet     & $120 \times 120$ & 43 & S2 (10) + S1 (2) & 20000     & 1000      & 1000      \\
x-so2sat          & $32 \times 32$   & 17 & S2 (10) + S1 (2) & 19992     & 986       & 986       \\
x-brick-kiln      & $64 \times 64$   & 2  & S2 (10) + S1 (2) & 15063     & 999       & 999       \\
x-eurosat         & $64 \times 64$   & 10 & S2 (10) + S1 (2) & 2000      & 1000      & 1000      \\
\midrule[\heavyrulewidth]
\multicolumn{7}{c}{\textit{Semantic Segmentation}} \\
\midrule
x-cashew-plantation & $256 \times 256$ & 7  & S2 (10) + S1 (2) & 1350      & 400       & 50        \\
x-SA-crop-type    & $256 \times 256$ & 10 & S2 (10) + S1 (2) & 3000      & 1000      & 1000      \\
x-harvey-building \new & $256 \times 256$ & 2  & S2 (10) + S1 (2) & 375       & 94        & 461       \\
x-sen1floods11 \new & $512 \times 512$ & 2  & S2 (10) + S1 (2) & 252       & 89        & 90        \\
\midrule[\heavyrulewidth]
\multicolumn{7}{c}{\textit{Change Detection}} \\
\midrule
x-harvey-flood \new & $256 \times 256$ & 2  & S2 (10) + S1 (2) & 375       & 94        & 461       \\
x-oscd \new        & $224 \times 224$ & 2  & S2 (10) + S1 (2) & 12 cities & 12 cities & 10 cities \\
\bottomrule
\end{tabular}
}
\end{table}

\paragraph{Bringing Sentinel-1 data.}
For the OSCD dataset \citep{oscd}, we combined it with the corresponding Sentinel-1 data collected by another team \citep{oscd-s1} to create x-oscd.
The m-so2sat dataset \citep{geobench, so2sat} from GeoBench already includes paired Sentinel-1 data with real and imaginary components for VH and VV. To create x-so2sat, we convert these components to log-power values in decibels:
\[
I_{p,\mathrm{dB}} = 10\log_{10}\!\left(s_{p,r}^{2}+s_{p,i}^{2}+\varepsilon\right),
\qquad p\in\{\mathrm{VV},\mathrm{VH}\},\quad \varepsilon=10^{-10},
\]
where $s_{p,r}$ and $s_{p,i}$ are the real and imaginary components supplied by So2Sat. This transformation computes squared magnitude followed by a logarithm; it does not compute the absolute magnitude of the complex signal.
For m-eurosat \citep{geobench}, we retrieve the corresponding Sentinel-1 data from EuroSAT-SAR \citep{fgmae} to create  x-eurosat. We create x-bigearthnet by pairing m-bigearthnet images \citep{geobench} with those from the original set \citep{bigearthnet} whose Sentinel-2 parts match those in GeoBench, and then retrieve the corresponding Sentinel-1 images.
We create x-cashew-plantation by pairing m-cashew-plantation \citep{geobench} with the corresponding Sentinel-1 images retrieved from the Copernicus Open Access Hub \citep{copernicus} using the dates provided in the Sentinel-2 version of the original data. We created x-SA-crop-type from the m-SA-crop-type \citep{geobench}.
The original set \citep{sa_crop_type} contains temporally close Sentinel-1 and Sentinel-2 image pairs, where each Sentinel-1 image was selected as the closest available in time to its corresponding Sentinel-2 image.
In m-SA-crop-type, 100 Sentinel-2 images were rotated.
To establish accurate matches between Sentinel-1 and Sentinel-2 images, we replaced these rotated images with nearest images from the original dataset.

The m-brick-kiln dataset proved to be the most challenging one, as the optical inputs \citep{geobench, brick-kiln} do not have individual acquisition dates: they are spatial crops of a Sentinel-2 temporal-mean composite constructed from acquisitions between October 2018 and May 2019. To construct x-brick-kiln, we query Sentinel-1 RTC VV/VH acquisitions that cover the same footprint during this interval. We sample up to 12 candidates across the date range and orbit directions, reproject each candidate to the north-up $64\!\times\!64$ optical grid, and reject crops with less than 95\% valid coverage or 90\% usable backscatter. Among the remaining candidates, we first maximize valid coverage and then the Pearson correlation between Sobel edge magnitudes computed from optical RGB--NIR and SAR VV--VH averages. We manually verified 100 randomly chosen pairs and did not find any visible issue with the matches.

\subsection{Evaluation Protocol}

GeoCrossBench evaluates models under three distinct settings designed to probe different aspects of generalization. For all settings, models are fine-tuned on the training data and then evaluated on the test data from various downstream datasets, each representing one of the three remote sensing tasks.
\paragraph{Setting 1: In-Distribution.}
This setting establishes a baseline performance metric. Models are trained and evaluated on the same set of bands. This measures the model's effectiveness in a standard, non-generalization setting. We test two common configurations:
\begin{itemize}[leftmargin=20pt, nosep]
    \item \textbf{Train on RGB $\rightarrow$ Evaluate on RGB:} Models are fine-tuned using only Sentinel-2's RGB bands (B4, B3, B2).
    \item \textbf{Train on S2 $\rightarrow$ Evaluate on S2:} Models are fine-tuned using all 10 available Sentinel-2 bands.
\end{itemize}

\paragraph{Setting 2: No-Overlap Bands.}
This setting tests a model's ability to transfer learned representations to a completely different sensor type, representing a challenging zero-shot generalization task.
\begin{itemize}[leftmargin=20pt, nosep]
    \item \textbf{Train on RGB $\rightarrow$ Evaluate on S1:} Generalization from RGB to SAR.
    \item \textbf{Train on S2 $\rightarrow$ Evaluate on S1:} Generalization from multispectral optical to SAR.
    \item \textbf{Train on RGB $\rightarrow$ Evaluate on N'S\textsubscript{1}S\textsubscript{2}:} Generalization from RGB to narrow near infrared, shortwave infrared 1 and 2 bands (S2 B8A, B11, B12).
\end{itemize}

\paragraph{Setting 3: Superset Bands.}
This setting assesses a model's robustness and ability to leverage new information when presented with more spectral bands at test time than it was trained on. This simulates a real-world scenario where a model trained on legacy data must operate on data from a newer, more capable sensor.
\begin{itemize}[leftmargin=20pt, nosep]
    \item \textbf{Train on RGB $\rightarrow$ Evaluate on RGBN:} Tests generalization from 3-band optical to 4-band optical, adding the Near-Infrared band (S2 B8).
    \item \textbf{Train on S2 $\rightarrow$ Evaluate on S2+S1:} Tests generalization from 10-band multispectral to a 12-band fused optical-SAR product.
\end{itemize}

\paragraph{Scene Classification.}
For scene classification tasks, models are trained to assign a class label to an entire image patch. The testing is done by using the evaluation band combination as input and the performance is measured using the corresponding evaluation metric for each task:
    \textbf{F1Score} for x-bigearthnet,
    \textbf{Accuracy} for x-so2sat, x-eurosat and x-brick-kiln.

\paragraph{Semantic Segmentation.}
For semantic segmentation, models are trained to assign a class label to each pixel in an image. At test time, the model segments images using the corresponding band combinations. Segmentation quality is measured by:
    \textbf{mIOU} for x-cashew-plantation, x-SA-crop-type, and x-sen1floods11, and
    \textbf{bIOU} for x-harvey-building.

\paragraph{Change Detection.}
Change detection tasks require the model to identify differences between two remote sensing images of the same area taken at different times. The evaluation trains on image pairs with the same band combinations and tests on pairs where the bands of the ``after'' image are replaced, while the bands of the ``before'' image remain unchanged. We report \textbf{bIOU} for x-harvey-flood and \textbf{F1Score} for x-oscd.

\section{Evaluated Models and Fine-Tuning Settings}

We evaluate a wide variety of models in two primary settings: (i) \textbf{full fine-tuning}, where all parameters of the pretrained foundation model and the task-specific head are updated; and (ii) \textbf{fine-tuning with frozen backbone}, where only the parameters of a newly added task-specific head (e.g., a linear layer for classification, a decoder for segmentation/change detection) are trained. These settings represent a trade-off between a model's adaptation capacity and preservation of the generalization capabilities that might come from pretraining.

\paragraph{Adapting inputs across band settings.}
When a pretrained model needs a wider input projection for downstream fine-tuning, we average the pretrained weights across input channels and use this mean, divided by the new channel count, for every channel in the wider projection.

At evaluation, we retain the downstream checkpoint's existing channel weights. Inputs with fewer bands than the projection expects are padded with zeros. For example, RGB$\rightarrow$SAR uses two SAR channels and one zero channel. If the input has more bands than the projection accepts, we initialize only the added channels with the mean of the existing weights, as in RGB$\rightarrow$RGBN. Neither evaluation operation involves further training.

\newmodel{} selects stored channel embeddings, DOFA receives wavelength metadata, and Panopticon receives optical-wavelength or SAR-polarization identifiers. Appendix~\ref{sec:input-band-adaptation} gives the adaptation details, and Section~\ref{sec:chivit-unseen-bands} explains how \newmodel{} handles bands absent during downstream fine-tuning.

\subsection{Pre-trained Foundation Models and Supervised Models}

\paragraph{Specialized Remote Sensing Foundation Models.}
We restrict the comparison to backbones with fewer than 100M parameters to keep the full benchmark feasible across many datasets, band settings, seeds, and hyperparameter searches, while retaining models suitable for memory- and latency-constrained downstream deployment. We picked several publicly available models pre-trained on remote sensing data in this range (ViT-B and Swin-B), namely TerraMind \cite{terramind}, TerraFM \cite{terrafm}, DOFA \cite{dofa}, SatlasNet \citep{satlas}, CROMA \cite{croma}, AnySat \cite{anysat}, Panopticon \citep{panopticon} and
Prithvi \cite{prithvi}.
The details of adapting the bands we need in the benchmarks to the expected inputs of the models are in Appendix \ref{fine-tune}.

\paragraph{General-purpose Image Foundation Models.}
We also include several general-purpose vision models as baselines.
We took self-supervised models of self-distillation type iBOT~\citep{ibot}, which is pretrained on ImageNet, DINOv2~\citep{dino_v2} and DINOv3~\citep{dino_v3} models pretrained on a huge custom dataset of images.
Because we use models with fewer than 100M parameters, only the ViT-B versions of DINOv2 and DINOv3 are used.
Moreover, models like the satellite version of DINOv3~\citep{dino_v3} or DINO-MC~\citep{dino_mc} lack the ViT-B version and they are not included in our comparison.
Following \cite{geobench}, we also evaluate ImageNet-pretrained ResNet-50 and ViT-B without self-supervised pre-training.
Recent work \citep{sup_baselines} has demonstrated that even non-pretrained models can produce competitive results with enough hyperparameter tuning.
We omit such baselines as we prefer fine-tuning recipes that are relatively easy and quick to implement for each new downstream task.

\paragraph{Hyperparameters.}
We perform a separate hyperparameter grid search for every dataset--model--training-band--adaptation-regime combination. Classification searches evaluate eight learning rates, while segmentation and change detection searches additionally evaluate three UPerNet widths, yielding 24 configurations per combination; Appendix \ref{hyperparam} specifies the grids. For segmentation and change detection tasks, we use the UPerNet head, which takes internal representations from four encoder layers; for change detection, we compute the difference between the representations of the two input images. Input sizes match those used during pretraining of the underlying model (following \citep{revisitingbenchmarks}).

\subsection{\newmodel{}: Self-supervised Channel-ViT on Remote Sensing Data}

\begin{figure}[t]
 \centering
 \includegraphics[clip, trim=0.0cm 0.5cm 0cm 0cm, width=\textwidth]{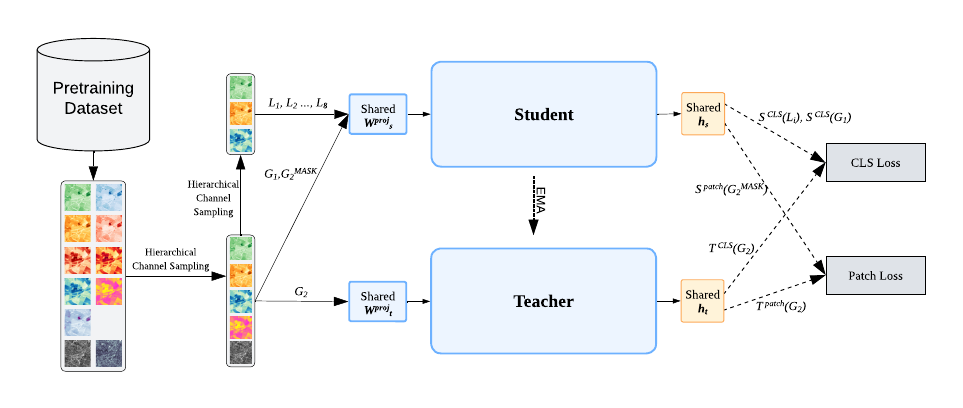}
 \caption{Overview of the iBOT-style self-distillation pretraining used for \newmodel{}. Hierarchical channel sampling is applied to create distinct views for the student and teacher, where student channels are a subset of the teacher's channels. Shared projection weights and a shared prediction head are utilized, with losses computed for both CLS and patch tokens.}
 \label{fig:chivit}
\end{figure}

Robust cross-band generalization requires learning transferable representations from partial spectral inputs. Motivated by recent advances in multi--channel self--supervision we extend ChannelViT~\citep{channelvit} with a hierarchical pre--training recipe tailored to remote--sensing imagery that we name \newmodel{} (ChiViT). The core idea is to give each spectral band equal importance during pretraining enabling the network to
a) fine-tuned on any band subset available without architectural changes and b) exchange information between spectrally distinct modalities.

\paragraph{Architecture.}
ChannelViT preserves channel-specific information by tokenizing each single band independently and adding a learnable \emph{channel embedding} $\mathbf{e}^{\mathrm{chn}}_{c}$ that is analogous to the positional embedding $\mathbf{e}^{\mathrm{pos}}{(h,w)}$ of ViT. Given an input $\mathbf{x}\in \mathbb{R}^{C\times H\times W}$, we partition every band into $N_p = (H/P) \cdot (W/P)$ non-overlapping patches of size $P \times P$. Unlike standard ViTs that create a single token from a multi-channel patch, ChannelViT generates one token from each single-channel patch. Thus, for each channel $c \in \{1, ..., C\}$ and each spatial patch $j \in \{1, ..., N_p\}$, we obtain a patch $x_{c,j}$. Each such single-channel patch is flattened into a vector of dimension $P^2$. These flattened patches are then linearly projected into $D$-dimensional embeddings using a learnable linear projection $W \in \mathbb{R}^{P^2 \times D}$. These projection weights $W$ are shared across all channels, promoting the learning of shared low-level features and enhancing robustness \citep{channelvit}.
To retain spatial and channel-specific information, learnable positional embeddings $\mathbf{e}^{\mathrm{pos}}_j \in \mathbb{R}^D$ (shared across channels) and learnable channel embeddings $\mathbf{e}^{\mathrm{chn}}_c \in \mathbb{R}^D$ are added to each projected patch token. A learnable classification token, $\mathbf{e}^{\mathrm{CLS}} \in \mathbb{R}^D$, is prepended to the sequence. The final sequence of $N = C \cdot N_p + 1$ tokens fed to the Transformer encoder is structured as: $[\mathbf{e}^{\mathrm{CLS}}; \dots; Wx_{c,j} + \mathbf{e}^{\mathrm{pos}}_j + \mathbf{e}^{\mathrm{chn}}_c; \dots]$. This allows the model to reason across both spatial locations and spectral channels simultaneously.

\paragraph{Pretraining dataset.}
To pretrain \newmodel{} for strong cross-band generalization, we extended Satlas Pretrain dataset \citep{satlas} up to over 23 million images. This dataset was collected to expose the model to a wide spectrum of Earth's surface characteristics, captured by various spectral bands and resolutions. Notably we added ``parallel'' data: the BigEarthNet \citep{bigearthnet} and Sen12MS datasets \citep{sen12ms}, offer Sentinel-1 and Sentinel-2 image pairs that are lined up, crucial for learning joint radar-optical features. While such parallel pairs were used in OlmoEarth,  Please refer to Appendix \ref{sec:pretraining} for all details.

Appendix \ref{pretraining_details} reports our search over training details.

\paragraph{Bands absent during fine-tuning.}\label{sec:chivit-unseen-bands}
Before downstream fine-tuning, \newmodel{} loads all 12 channel embeddings from the pretrained teacher checkpoint. Channel IDs 0--2 correspond to RGB, IDs 0--9 to the ten S2 bands, and IDs 10--11 to the SAR polarizations VV and VH. During full fine-tuning, the shared patch projection, spatial positional embeddings, and Transformer are updated, while only channel embeddings selected by the training band IDs receive task-loss gradients. Embeddings for absent bands remain stored and receive no task-loss gradients. With a frozen backbone, all channel embeddings remain fixed. At test time, the input band IDs select the corresponding embeddings from the downstream checkpoint, including those for bands absent during downstream training. No channel embeddings are reinitialized, and no test-time optimization is performed. Thus, for \newmodel{}, downstream-unseen bands are drawn from its pretrained channel vocabulary.

\section{GeoCrossBench Experiments and Discussion}\label{sec:experiments}
\begin{table}[b!]
\centering
\caption{Performance evaluation of all tested models on GeoCrossBench. Each direct-transfer column averages the designated bounded metric over the benchmark datasets. Setting averages combine their constituent transfer columns, while Overall AVG is the mean of the three setting averages. An asterisk indicates a frozen backbone; higher is better.}
\label{tab:ranking}
\includegraphics[width=\linewidth]{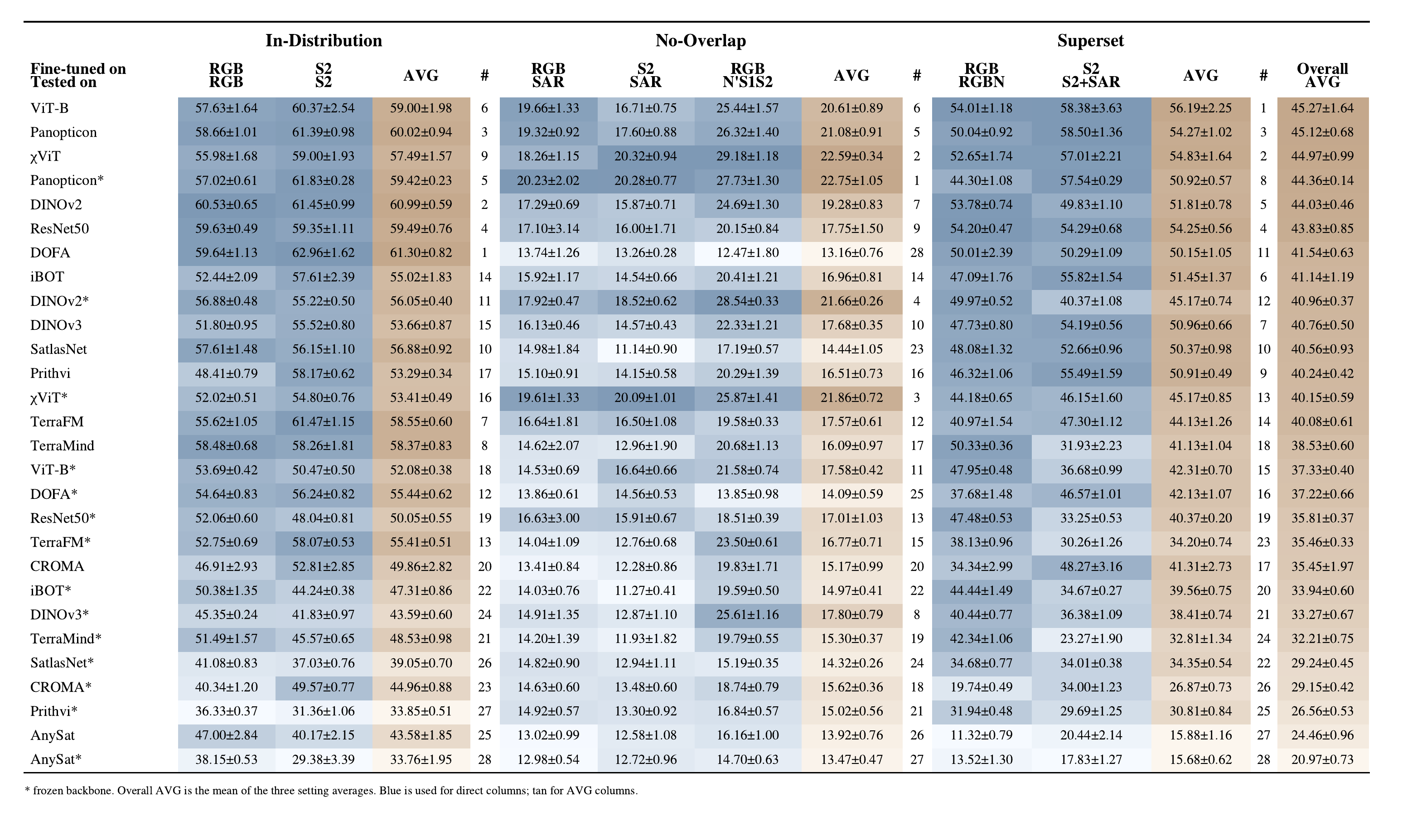}
\end{table}

Table \ref{tab:ranking} shows the ranking of all tested models in each evaluation setting. For each seed, we first aggregate across datasets within each transfer and setting; the reported standard deviation is then computed across the seed-level aggregates. 
Appendix~\ref{app:pairwise-significance} reports pairwise statistical tests of Overall scores using paired seed--dataset bootstrap confidence intervals.

\paragraph{Model rankings depend on the transfer setting.}
In the \emph{In-Distribution} setting, DOFA has the highest average (61.30), followed by DINOv2 (60.99) and Panopticon (60.02). Thus, general-purpose vision pretraining remains highly competitive with remote-sensing-specific pretraining when training and test bands match.

\paragraph{RS foundation models are limited in their cross-band generalization.}
No-overlap transfer remains the most difficult setting. Frozen Panopticon has the highest no-overlap average (22.75), followed closely by \newmodel{} (22.59) and frozen \newmodel{} (21.86)\footnote{A post-hoc audit identified pretraining--evaluation overlap affecting BigEarthNet, which we do not expect to materially alter the paper's overall conclusions. Nearly all images in the GeoBench BigEarthNet test split occurred, without labels, in the self-supervised pretraining corpus of \newmodel{}. Pretraining preceded our alignment of the downstream task with the original GeoBench split, and we did not retrain \newmodel{} because no BigEarthNet labels were used. Separately, CROMA's original hyperparameters were selected partly using labeled BigEarthNet performance.}. The top two differ by only 0.16 points on average, and their ordering changes across seeds.

In the \emph{Superset} setting, ViT-B has the highest average (56.19), followed by \newmodel{} (54.83) and Panopticon (54.27). At the aggregate level, every tested model performs worse for both superset transfers than for the corresponding in-distribution transfer, and this pattern holds for every seed-level aggregate.

\paragraph{Frozen backbones can remain competitive.} Full fine-tuning is often stronger, but frozen variants are especially competitive in the \textit{No-Overlap} setting: frozen Panopticon and frozen \newmodel{} rank first and third, respectively (Table \ref{tab:ranking}).

\paragraph{The value of RS-specific pretraining.} One can ask what additional knowledge can RS-specific foundation models learn that will help them beat general-purpose models. The good performance of Panopticon and \newmodel{} in the \textit{No-Overlap} setting suggests that carefully mixing imagery with varied spectral coverage and modalities during pretraining may help models learn cross-band relationships. But even in this setting, DINOv2 has the 4th best result, above most RS-specific models. When we take the average of the three evaluation settings, three of the top six models are general-purpose models.

\begin{figure}[t]
    \centering
    \includegraphics[width=\textwidth]{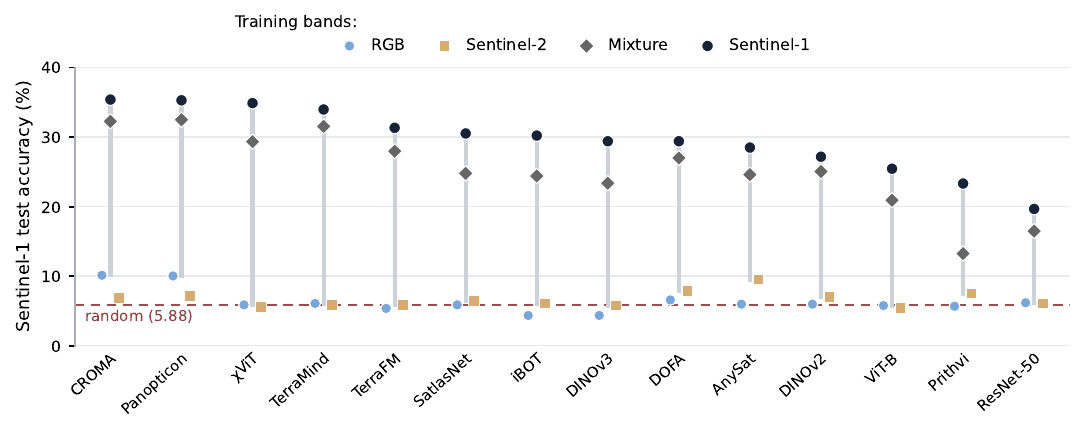}
    \caption{Frozen linear-probe accuracy on the x-so2sat Sentinel-1 test set when trained on various band combinations. The dashed line marks the random baseline for 17 classes.}
    \label{fig:so2sat-saturation}
\end{figure}

\paragraph{Is the benchmark saturated?} One way to probe whether the benchmark is saturated is to measure whether models perform better when oracle-labeled data is available for the target bands. On x-so2sat, we train linear probes on RGB, Sentinel-2, Sentinel-1, or a size-matched mixture of RGB, Sentinel-2, Sentinel-1, and N'S1S2 representations, and evaluate every probe on the Sentinel-1 test set. Each training regime contains the same number of labeled examples, and we report the mean over five random assignments. As shown in Figure~\ref{fig:so2sat-saturation}, current optical-to-SAR transfers are generally near the 5.88\% random baseline. In contrast, direct Sentinel-1 supervision raises accuracy to ~35\% for CROMA, Panopticon and \newmodel{}, demonstrating substantial headroom for improvement. Mixture training also improves over optical-only training, but remains below Sentinel-1-only training possibly because of the train-test shift in the band distribution.

\paragraph{Implications for future models.} One of the reasons for the relatively strong performance of Panopticon and \newmodel{} compared to other multispectral models might be the trick of \textit{sampling of the bands} during pretraining. The models might learn to rely less on band-specific features and instead focus on patterns shared across bands, which then improves cross-band generalization performance. We also tried applying Hierarchical Channel Sampling (HCS) during fine-tuning. Contrary to previous findings \citep{channelvit}, our experiments on a subset of datasets (see Appendix \ref{sec:ft_sampling}) show inconsistent improvements. While HCS aids generalization in specific scenarios, it does not universally improve performance.

The impact of the \textit{scale of the models and datasets} is relatively underexplored in remote sensing. Larger variants do not uniformly dominate smaller ones across remote-sensing tasks, as also observed for Galileo and OlmoEarth \citep{galileo,olmoearth}. GeoCrossBench limits model size during inference, but larger models may still be useful as teachers for distillation, as demonstrated in DINOv2.

Finally, the \textit{quality and quantity of remote-sensing pretraining data} can affect performance on benchmarks such as GeoCrossBench.
While co-located optical-SAR imagery is already used by several remote-sensing foundation models \citep{croma,decur,mmearth,anysat,panopticon,galileo,terrafm,terramind,olmoearth}, we believe its role should significantly increase in future models that prioritize cross-satellite knowledge transfer.
A promising direction is to study how explicitly paired samples are selected and used in pretraining objectives, and to discover more sources of parallel imagery.

\section{Related Work}
The development of foundation models for Earth observation has seen rapid progress, aiming to create versatile models applicable to a wide array of downstream tasks.
Their versatility is measured by transfer to these downstream tasks, that is, by using them on different data across geography, time, and modalities.

\paragraph{Foundation Models for Remote Sensing.}
Remote sensing data is in effect its own data modality, with unique challenges and opportunities, and so there has been a call for specialized machine learning approaches and models \citep{missioncritical}.
This large-scale/small-scale tension motivates foundation modeling to enable more efficient transfer and application across tasks with limited labels.
We highlight pioneering methods that established the topic: SatMAE \citep{satmae} demonstrated self-supervised pre-training for RS at ViT scale and Satlas \citep{satlas} by contrast demonstrated large-scale and multi-task supervised pre-training for RS.
Scale-MAE \citep{scalemae} refined self-supervised learning for RS by focusing on different spatial resolutions and generalization across them by controlling for ground sample distance (i.e. the physical size of a pixel).
Our \newmodel{} model and the GeoCrossBench dataset are related in spirit to Scale-MAE but spectral, rather than spatial, and explores how to generalize across bands instead of resolutions.
GeoCrossBench is a response to the call to do machine learning for remote sensing: it measures the specific need to generalize across bands given the variation in spectral coverage across satellite sensors and the uneven availability of labeled data for each band configuration.

\paragraph{Multi-modal/sensor/band Learning in Remote Sensing.}
Multi-modal data with many different bands is common in remote sensing because satellite sensors vary in spectral coverage and sensing modality.
As in self-supervised deep learning for other modalities, foundation models in RS learn from multi-modal data in RS:
SatMAE \citep{satmae}, Scale-MAE \citep{scalemae}, and MMEarth \citep{mmearth} auto-encode multispectral optical data and MMEarth decodes other modalities, while TerraMind \citep{terramind} enables any-to-any generation across pixel and token level modalities;
SoftCon \citep{softcon}, DeCUR \citep{decur}, and DOFA \citep{dofa} separately learn intra-modal representations of multi-spectral and radar data; and
CROMA \citep{croma}, AnySat \citep{anysat}, TerraFM \citep{terrafm} and Galileo \citep{galileo} jointly learn inter-modal representations of multi-spectral and radar data (CROMA) and more modalities like elevation or climate (AnySat, Galileo, TerraMind).
In summary these works explore many ways to learn \emph{from} bands but not \emph{across} bands and do not cover how to extend or generalize to new or different bands.
GeoCrossBench highlights this direction of improvement, and measures the need for improvement, which is practically motivated by the cost to (re-)train these ever larger foundation models.
It is not feasible to train for all combinations of bands, at least not for most groups, so generalization is necessary.

\paragraph{Datasets and Benchmarking for Remote Sensing.}
Shared datasets and benchmarks are key for comparability in the context of the diversity of RS data and tasks.
Our focus is evaluation, like GEO-Bench \citep{geobench}, rather than pre-training data, such as Terra \citep{terradata} and SSL4EO-S12 \citep{ssl4eo}.
There are many and high-quality task-specific benchmarks (for marine debris \citep{kikaki2024detecting}, floods \citep{bonafilia2020sen1floods11}, agriculture \citep{garnot2021panoptic,russwurm2019breizhcrops,tseng2021cropharvest}, and more) but they do not focus on general capacities like generalization or efficiency.
GeoCrossBench is needed because no existing benchmark measures our key question of cross-band generalization.

\section*{Limitations and Conclusion} We focus on evaluating the cross-band generalization of remote sensing foundation models by investigating within optical bands and across optical and radar bands by extending existing datasets with SAR data. The datasets are limited to static objects and scenes and do not cover moving objects for which it is extremely hard to find parallel optical-SAR imagery. Even if the models achieve perfect scores on our benchmark, they might struggle in detecting moving objects on unseen bands.
Our aggregate results show setting-dependent leaders: DOFA in-distribution, frozen Panopticon for no-overlap transfer, and ViT-B for superset transfer and overall. The top three overall averages differ by only 0.30 points, so we do not claim a decisive overall winner. Across all models, the decline from in-distribution to the superset setting is substantial, while the no-overlap setting produces a much larger drop. General-purpose models remain competitive with remote-sensing-specific models. Nevertheless, the leading no-overlap configurations suggest that pretraining with co-located observations from multiple sensors may offer remote-sensing-specific models a path toward a clearer advantage. We hope GeoCrossBench motivates more robust spectral-transfer methods and more complete reporting of metrics, uncertainty, and tuning protocols.

We have used ChatGPT and Gemini to polish the writing. We have used Codex to identify issues in our large grid of fine-tuning and evaluation runs, rerun failed jobs, and prepare the tables.
\bibliographystyle{tmlr}

\bibliography{geocrossbench}

\newpage
\appendix

\section{Pairwise Statistical Comparisons}
\label{app:pairwise-significance}
To assess pairwise ranking stability, we use a paired crossed bootstrap over training seeds and benchmark datasets. We generate 50,000 replicates; in each replicate, the same five seed indices are resampled with replacement for every model, and datasets are resampled with replacement within task type (four classification, four segmentation, and two change-detection datasets), with these shared draws preserving pairwise alignment. We recompute each model's three setting averages and Overall score, then form row-minus-column differences. Table~\ref{tab:pairwise-overall-bootstrap} reports the observed differences and pointwise percentile 95\% confidence intervals. These intervals do not adjust for multiple comparisons and do not include test-sample uncertainty.

The six highest-ranked configurations (ViT-B, Panopticon, \newmodel{}, frozen Panopticon, DINOv2, and ResNet50) are mutually statistically indistinguishable under this analysis. DOFA, the next-ranked configuration, is significantly worse than 5 of the six; its comparison with frozen Panopticon remains inconclusive.

\begin{table}[h!]
\centering
\caption{Pairwise comparisons among the 12 highest-ranked model configurations by Overall GeoCrossBench score. Each lower-triangle cell reports the observed row-minus-column difference and its pointwise 95\% paired crossed-bootstrap confidence interval in smaller type. A shaded background marks an interval that excludes zero. An asterisk indicates a frozen backbone.}
\label{tab:pairwise-overall-bootstrap}
\begingroup
\newcommand{\paircell}[2]{\shortstack{\fontsize{7.2}{7.6}\selectfont $\mathbf{#1}$\\[-0.7mm]\fontsize{4.0}{4.3}\selectfont $#2$}}
\setlength{\tabcolsep}{1.0pt}
\renewcommand{\arraystretch}{1.20}
\resizebox{\textwidth}{!}{%
\begin{tabular}{@{}l*{12}{c}@{}}
\toprule
 & \rotatebox{60}{\strut ViT-B} & \rotatebox{60}{\strut Panopticon} & \rotatebox{60}{\strut \newmodel{}} & \rotatebox{60}{\strut Panopticon*} & \rotatebox{60}{\strut DINOv2} & \rotatebox{60}{\strut ResNet50} & \rotatebox{60}{\strut DOFA} & \rotatebox{60}{\strut iBOT} & \rotatebox{60}{\strut DINOv2*} & \rotatebox{60}{\strut DINOv3} & \rotatebox{60}{\strut SatlasNet} & \rotatebox{60}{\strut Prithvi} \\
\midrule
\textbf{ViT-B} & $\cdot$ &  &  &  &  &  &  &  &  &  &  &  \\
\textbf{Panopticon} & \paircell{-0.14}{[-2.76,+3.18]} & $\cdot$ &  &  &  &  &  &  &  &  &  &  \\
\textbf{\newmodel{}} & \paircell{-0.30}{[-4.18,+4.16]} & \paircell{-0.15}{[-2.66,+2.16]} & $\cdot$ &  &  &  &  &  &  &  &  &  \\
\textbf{Panopticon*} & \paircell{-0.90}{[-5.62,+4.12]} & \paircell{-0.76}{[-4.09,+2.59]} & \paircell{-0.61}{[-3.49,+2.18]} & $\cdot$ &  &  &  &  &  &  &  &  \\
\textbf{DINOv2} & \paircell{-1.24}{[-4.39,+2.20]} & \paircell{-1.10}{[-4.79,+2.59]} & \paircell{-0.94}{[-4.79,+2.33]} & \paircell{-0.34}{[-3.65,+2.89]} & $\cdot$ &  &  &  &  &  &  &  \\
\textbf{ResNet50} & \paircell{-1.44}{[-4.99,+2.37]} & \paircell{-1.29}{[-4.67,+1.96]} & \paircell{-1.14}{[-3.88,+1.06]} & \paircell{-0.53}{[-2.98,+1.82]} & \paircell{-0.20}{[-2.62,+2.55]} & $\cdot$ &  &  &  &  &  &  \\
\textbf{DOFA} & \cellcolor{pairwisesignificant}\paircell{-3.73}{[-6.48,-0.66]} & \cellcolor{pairwisesignificant}\paircell{-3.59}{[-6.47,-0.70]} & \cellcolor{pairwisesignificant}\paircell{-3.43}{[-6.66,-0.74]} & \paircell{-2.83}{[-5.84,+0.25]} & \cellcolor{pairwisesignificant}\paircell{-2.49}{[-4.20,-0.70]} & \cellcolor{pairwisesignificant}\paircell{-2.29}{[-4.41,-0.38]} & $\cdot$ &  &  &  &  &  \\
\textbf{iBOT} & \cellcolor{pairwisesignificant}\paircell{-4.12}{[-5.53,-2.83]} & \cellcolor{pairwisesignificant}\paircell{-3.98}{[-7.73,-1.27]} & \paircell{-3.83}{[-8.63,+0.11]} & \paircell{-3.22}{[-8.25,+1.39]} & \cellcolor{pairwisesignificant}\paircell{-2.88}{[-5.95,-0.13]} & \paircell{-2.68}{[-6.56,+0.73]} & \paircell{-0.39}{[-3.32,+1.96]} & $\cdot$ &  &  &  &  \\
\textbf{DINOv2*} & \cellcolor{pairwisesignificant}\paircell{-4.31}{[-7.47,-0.91]} & \cellcolor{pairwisesignificant}\paircell{-4.16}{[-7.39,-1.04]} & \cellcolor{pairwisesignificant}\paircell{-4.01}{[-7.08,-1.31]} & \cellcolor{pairwisesignificant}\paircell{-3.40}{[-5.87,-0.97]} & \cellcolor{pairwisesignificant}\paircell{-3.07}{[-4.39,-1.60]} & \cellcolor{pairwisesignificant}\paircell{-2.87}{[-4.92,-1.13]} & \paircell{-0.58}{[-1.87,+0.86]} & \paircell{-0.19}{[-3.15,+2.97]} & $\cdot$ &  &  &  \\
\textbf{DINOv3} & \cellcolor{pairwisesignificant}\paircell{-4.50}{[-5.59,-3.15]} & \cellcolor{pairwisesignificant}\paircell{-4.36}{[-7.01,-1.96]} & \cellcolor{pairwisesignificant}\paircell{-4.20}{[-7.78,-0.82]} & \paircell{-3.60}{[-7.88,+0.66]} & \cellcolor{pairwisesignificant}\paircell{-3.26}{[-5.71,-0.80]} & \cellcolor{pairwisesignificant}\paircell{-3.06}{[-6.20,-0.04]} & \paircell{-0.77}{[-2.98,+1.36]} & \paircell{-0.38}{[-1.56,+1.19]} & \paircell{-0.19}{[-2.80,+2.42]} & $\cdot$ &  &  \\
\textbf{SatlasNet} & \paircell{-4.70}{[-8.90,+0.09]} & \cellcolor{pairwisesignificant}\paircell{-4.56}{[-8.31,-0.61]} & \cellcolor{pairwisesignificant}\paircell{-4.41}{[-7.79,-1.29]} & \cellcolor{pairwisesignificant}\paircell{-3.80}{[-6.10,-1.57]} & \cellcolor{pairwisesignificant}\paircell{-3.46}{[-6.50,-0.57]} & \cellcolor{pairwisesignificant}\paircell{-3.27}{[-4.57,-1.74]} & \paircell{-0.97}{[-3.37,+1.40]} & \paircell{-0.58}{[-4.37,+3.90]} & \paircell{-0.40}{[-2.66,+2.12]} & \paircell{-0.20}{[-3.90,+3.77]} & $\cdot$ &  \\
\textbf{Prithvi} & \cellcolor{pairwisesignificant}\paircell{-5.03}{[-8.11,-2.22]} & \cellcolor{pairwisesignificant}\paircell{-4.89}{[-9.94,-0.28]} & \paircell{-4.73}{[-10.48,+0.07]} & \paircell{-4.13}{[-10.07,+1.66]} & \cellcolor{pairwisesignificant}\paircell{-3.79}{[-6.84,-0.81]} & \paircell{-3.59}{[-8.12,+0.67]} & \paircell{-1.30}{[-4.59,+1.90]} & \paircell{-0.91}{[-3.47,+1.46]} & \paircell{-0.72}{[-4.53,+2.94]} & \paircell{-0.53}{[-3.25,+1.93]} & \paircell{-0.33}{[-5.16,+4.55]} & $\cdot$ \\
\bottomrule
\end{tabular}%
}
\endgroup
\end{table}

\section{Frozen Backbones vs. Full Fine-Tuning} \label{frozen}

\begin{figure}[htbp]
    \centering
    \includegraphics[width=\textwidth]{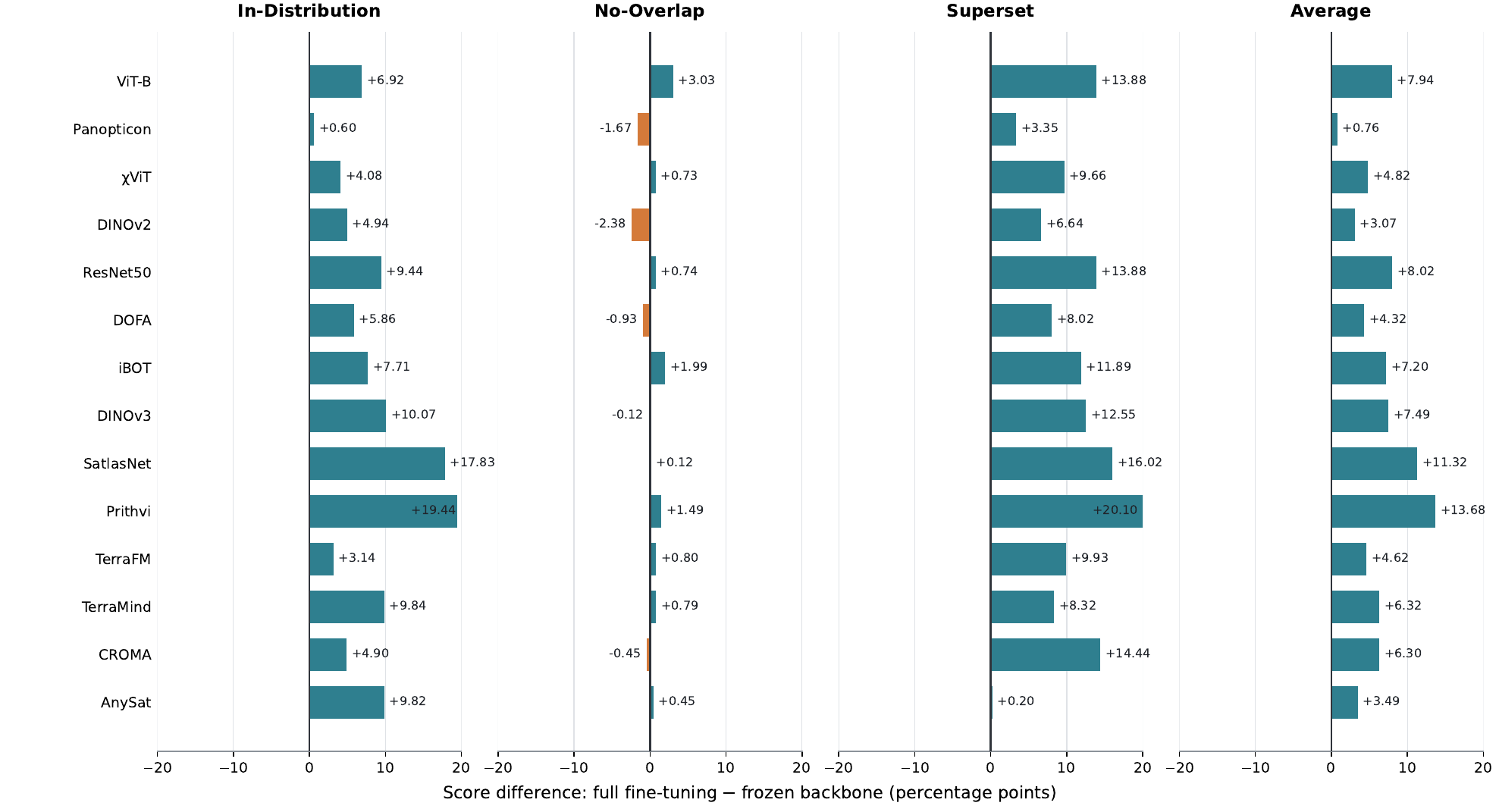}
    \caption{Difference in aggregate score between full fine-tuning and frozen-backbone adaptation for each model and evaluation setting. Positive values favor full fine-tuning; negative values favor the frozen backbone. Average denotes the mean of the in-distribution, no-overlap, and superset scores.}
    \label{fig:frozen-vs-full}
\end{figure}

Figure~\ref{fig:frozen-vs-full} compares full fine-tuning with frozen-backbone adaptation for each model. Full fine-tuning improves every model in the in-distribution and superset settings, and consequently improves every model's average across the three settings. The no-overlap setting is different: the differences are smaller and mixed in direction, with frozen variants outperforming full fine-tuning for five of the fourteen models. One possible explanation is that the benefit of tuning the whole backbone for the task is lost because of catastrophic forgetting of the knowledge about the bands not seen in fine-tuning.

\section{Pretraining dataset} \label{sec:pretraining}
The main components are summarized in Table~\ref{tab:pretraining_dataset}.
A key part comes from the Satlas Pretrain dataset \citep{satlas}, which includes Sentinel-1 SAR images in VV and VH polarizations, Sentinel-2 multispectral images (using 10 bands, excluding B01, B09, and B10 which have lower resolution), and NAIP high-resolution RGB aerial photos.
The MillionAID dataset \citep{millionaid} contributes a large volume of RGB aerial images with varied resolutions and image sizes.
The BigEarthNet \citep{bigearthnet} and Sen12MS datasets \citep{sen12ms}, offer Sentinel-1 and Sentinel-2 image pairs that are lined up, crucial for learning joint radar-optical feature.
The Intelinair dataset \citep{intelinair} gives very detailed (0.02m GSD) RGB and Near-Infrared (NIR) aerial images of farms.
Using all these different datasets together gives \newmodel{} a solid base for pretraining. This helps the model learn features that work well across different kinds of sensors.

\begin{table}[h]
\centering
\caption{Overview of datasets used for \newmodel{} pretraining.}
\label{tab:pretraining_dataset}
\small
\begin{tabular}{@{}lllll@{}}
\toprule
Dataset & Bands & \# Images & GSD & Image Size \\ \midrule
Satlas \citep{satlas} (S1) & Sentinel-1 (VV, VH) & 4.5M & 20m & $512 \times 512$ \\
Satlas \citep{satlas} (S2) & Sentinel-2 (10 bands) & 11M & 10m & $512 \times 512$ \\
Satlas \citep{satlas} (NAIP) & Aerial RGB & 5.3M & 1m & $512 \times 512$ \\
MillionAID \citep{millionaid} & Aerial RGB & 2M & 0.5-153m & $(170-550)^2$ \\
BigEarthNet \citep{bigearthnet} & S1 SAR \& S2 (10 bands) & 0.55M & 10m & $120 \times 120$ \\
Sen12MS \citep{sen12ms} & S1 SAR \& S2 (10 bands) & 0.18M & 10m & $256 \times 256$ \\
Intelinair \citep{intelinair} & Aerial RGB, NIR & 34K & 0.02m & $320 \times 320$ \\ \bottomrule
\end{tabular}
\end{table}

\section{Pretraining details} \label{pretraining_details}
The pre-training process is visualized in Figure~\ref{fig:chivit}, where we utilized a multi-crop setup with 8 local views (denoted as $L_i$ in the figure) and 2 global views (e.g., $G_1, G_2$).
For the final pretraining of \newmodel{}, we processed 400 million samples in total. The AdamW optimizer \citep{adamw} was used with a batch size of 512. Given that the final pretraining did not have a predefined number of iteration steps, as we aimed to train for as long as we could manage, we utilized the Warmup-Stable-Decay (WSD) learning rate scheduler \citep{wsd}.  This approach allowed for a flexible decay phase, which was initiated for the last 10\% of total iterations. The learning rate was linearly warmed up for the initial 30 million samples to a peak of $2.5 \times 10^{-4}$, maintained during the stable phase, before the final decay. It's common practice to adjust this peak learning rate in proportion to the batch size (e.g., using the formula $peak\_lr \times batch\_size / 256$). The overall loss was a sum of the [CLS] token self-distillation loss and the MIM (masked image modeling) loss, without scaling factors between them.

\paragraph{Design choices.} To determine the optimal configuration for \newmodel{}, we conducted several experiments for the key design choices. Each experimental configuration was pre-trained for 40 million samples. Model selection was based on the mean Average Precision (mAP) achieved after fine-tuning on 1\% of the BigEarthNet dataset. The results of these ablation studies are summarized in Table~\ref{tab:design_choices_pretraining}.
Based on these experiments, the winning configuration utilized subset sampling for student channels by sampling from teacher channels (employing hierarchical channel sampling as described in \citep{channelvit}), shared projection weights for all bands, a shared prediction head for CLS and patch tokens, and a parallel data coefficient of 4 for the BigEarthNet and Sen12MS datasets during pretraining.
\begin{table}[h]
\centering
\caption{Ablation study of \newmodel{} pretraining choices. All configurations were pre-trained for 40M samples and evaluated by fine-tuning on 1\% of BigEarthNet using mAP. ``Subset Sampling'' draws the student's channels from those visible to the teacher; ``Shared Proj.'' shares one patch projection across bands; and ``Shared Head'' shares one prediction head across the CLS- and patch-token objectives. PDC denotes the Parallel Data Coefficient.}
\label{tab:design_choices_pretraining}
\begin{tabular}{@{}cccccc@{}}
\toprule
Subset Sampling & Shared Proj. & Shared Head & PDC ($\lambda$) & mAP (\%) \\ \midrule
\pmb{\checkmark} & \pmb{\checkmark} & \pmb{\checkmark} & \textbf{4} & \textbf{54.72} \\
x & \checkmark & \checkmark & 4 & 51.10 \\
\checkmark & x & \checkmark & 4 & 40.44 \\
\checkmark & \checkmark & x & 4 & 46.55 \\
\checkmark & \checkmark & \checkmark & 8 & 51.51 \\ \bottomrule
\end{tabular}
\end{table}

\section{Fine-tuning} \label{fine-tune}
\subsection{Input Bands Adaptation}\label{sec:input-band-adaptation}
When a pretrained model needs a wider input projection for downstream fine-tuning, we average its first-layer kernels across the original input channels. We replicate this average across the $C$ channels of the new projection and divide each copy by $C$. Here $C$ is the number of channels accepted by the model.

The input layer may expect more channels than the image provides. We place the selected bands in the first input channels, in the order returned by the data loader, and fill the remaining channels with zeros. For example, a model with a 12-channel input projection can be fine-tuned on ten S2 bands followed by two zero channels.

Evaluation starts from the downstream checkpoint. If the test image has fewer bands than its input projection expects, we append zero channels and keep the first-layer weights unchanged. Thus, an RGB-trained three-channel checkpoint receives two SAR bands followed by one zero channel. Replacing RGB with three non-overlapping optical bands changes the spectral information carried by the same input channels while leaving the projection unchanged.

If the test bands exceed the checkpoint's input width, we expand the projection while retaining its existing channel kernels and bias. Each added kernel is initialized with the mean of the checkpoint's existing kernels. For RGB$\rightarrow$RGBN, this adds a fourth kernel for NIR. A checkpoint already configured for 12 channels needs no expansion for S2$\rightarrow$S2+SAR: the two SAR bands replace the zero channels used during S2 fine-tuning. These evaluation operations are deterministic and do not involve optimization on test inputs.

\subsection{Hyperparameters}
\label{hyperparam}

We independently tune every dataset--model--training-band--adaptation-regime combination. For full fine-tuning, we search learning rates in $\{10^{-5}, 3{\times}10^{-5}, 5{\times}10^{-5}, 6{\times}10^{-5}, 10^{-4}, 3{\times}10^{-4}, 5{\times}10^{-4}, 6{\times}10^{-4}\}$. For frozen-backbone training, we search $\{10^{-4}, 3{\times}10^{-4}, 4{\times}10^{-4}, 5{\times}10^{-4}, 10^{-3}, 3{\times}10^{-3}, 4{\times}10^{-3}, 5{\times}10^{-3}\}$. For segmentation and change detection, we additionally search UPerNet width values in $\{1,2,3\}$. Thus, each classification search evaluates eight configurations, whereas each segmentation or change-detection search evaluates 24 configurations. All grid searches use seed 42.

\subsection{Fine-tuning Details}
Across all tasks, we consistently apply the following settings. We use a learning-rate scheduler featuring a 20-epoch linear warmup phase, which is then followed by a cosine decay. This decay period lasts for 30 epochs in classification tasks, and 80 epochs for both segmentation and change detection tasks. We use AdamW optimizer \citep{adamw} for all model trainings. For input normalization, we apply channel-wise mean and standard deviation normalization, clipping the resulting values to the range $[-3,3]$. We select the checkpoint with the highest validation metric. We set batch sizes to 64 for classification tasks, and to 8 for both segmentation and change detection tasks.

\section{Impact of Channel Sampling during Fine-Tuning}
\label{sec:ft_sampling}
To check if the benefits of Hierarchical Channel Sampling (HCS) apply to fine-tuning, we trained \newmodel{} with HCS on three datasets: x-bigearthnet (classification), x-SA-crop-type (segmentation), and x-harvey-flood (change detection).

The results are presented in Table \ref{tab:ft_sampling}. We observe that the impact of HCS is highly dataset-dependent. On x-bigearthnet, subsampling helps in the \textit{No-Overlap} setting, but on x-SA-crop-type and x-harvey-flood, the improvement is negligible or inconsistent.

\begin{table}[h]
\centering
\caption{Comparison of standard fine-tuning versus fine-tuning with Hierarchical Channel Sampling (HCS) using \newmodel{}. The table reports performance across all three evaluation settings.}
\label{tab:ft_sampling}
\resizebox{\textwidth}{!}{%
\begin{tabular}{@{}llccccccc@{}}
\toprule
 & & \multicolumn{2}{c}{\textbf{In-Distribution}} & \multicolumn{3}{c}{\textbf{No-Overlap}} & \multicolumn{2}{c}{\textbf{Superset}} \\
\cmidrule(lr){3-4} \cmidrule(lr){5-7} \cmidrule(lr){8-9}
\textbf{Dataset} & \textbf{Method} & \textbf{RGB$\to$RGB} & \textbf{S2$\to$S2} & \textbf{RGB$\to$S1} & \textbf{S2$\to$S1} & \textbf{RGB$\to$N'S\textsubscript{1}S\textsubscript{2}} & \textbf{RGB$\to$RGBN} & \textbf{S2$\to$S2+S1} \\ \midrule

x-bigearthnet
 & Standard & $71.03 \pm 0.88$ & $73.15 \pm 0.43$ & $5.14 \pm 1.29$ & $6.05 \pm 1.37$ & $19.34 \pm 3.42$ & $68.31 \pm 1.18$ & $70.91 \pm 0.55$ \\
 & HCS & $67.39 \pm 0.59$ & $70.85 \pm 0.49$ & $6.83 \pm 2.49$ & $16.95 \pm 4.72$ & $21.88 \pm 13.49$ & $66.12 \pm 0.79$ & $69.49 \pm 1.91$ \\ \midrule

x-SA-crop-type
 & Standard & $19.58 \pm 1.28$ & $30.85 \pm 0.54$ & $5.12 \pm 0.09$ & $5.09 \pm 0.02$ & $19.67 \pm 2.16$ & $20.08 \pm 1.17$ & $29.49 \pm 0.42$ \\
 & HCS & $24.04 \pm 0.91$ & $30.99 \pm 1.07$ & $5.13 \pm 0.07$ & $5.07 \pm 0.00$ & $23.10 \pm 0.57$ & $24.46 \pm 0.97$ & $29.49 \pm 0.97$ \\ \midrule

x-harvey-flood
 & Standard & $64.30 \pm 7.87$ & $56.22 \pm 12.07$ & $4.11 \pm 3.18$ & $10.43 \pm 4.80$ & $18.01 \pm 6.86$ & $59.24 \pm 6.08$ & $51.94 \pm 12.43$ \\
 & HCS & $70.66 \pm 0.94$ & $45.72 \pm 10.68$ & $10.59 \pm 6.40$ & $3.64 \pm 1.77$ & $46.90 \pm 2.47$ & $59.70 \pm 2.29$ & $40.98 \pm 8.08$ \\
\bottomrule
\end{tabular}%
}
\end{table}

\section{Wavelength Sensitivity} \label{sec:color_blind}

To investigate if models rely on specific spectral signatures or wavelength-independent spatial structures, we fine-tuned a ``color-blind'' DINOv3 where embedding weights are shared across all input bands. As shown in Table \ref{tab:color_blind}, while the color-blind model sees a drop in \textit{In-Distribution} performance (RGB$\to$RGB), it improves generalization to the distinct SAR modality (RGB$\to$S1). This suggests that enforcing wavelength insensitivity encourages the model to rely on geometric and textural features that are more robust to domain shifts.

\begin{table}[h]
\centering
\caption{Comparison of standard vs. ``color-blind'' DINOv3 (shared embedding weights). The color-blind model improves zero-shot transfer to SAR (S1) at the cost of in-distribution accuracy.}
\label{tab:color_blind}
\begin{tabular}{@{}llccc@{}}
\toprule
 & & \textbf{In-Dist.} & \multicolumn{2}{c}{\textbf{No-Overlap}} \\
\cmidrule(lr){3-3} \cmidrule(lr){4-5}
\textbf{Dataset} & \textbf{Model} & \textbf{RGB$\to$RGB} & \textbf{RGB$\to$S1} & \textbf{RGB$\to$N'S\textsubscript{1}S\textsubscript{2}} \\ \midrule

x-so2sat
 & DINOv3 & $60.83 \pm 1.66$ & $5.66 \pm 0.55$ & $11.03 \pm 1.84$ \\
 & color-blind DINOv3 & $54.32 \pm 2.74$ & $8.32 \pm 1.41$ & $17.26 \pm 2.51$ \\ \midrule

x-SA-crop-type
 & DINOv3 & $32.32 \pm 0.38$ & $5.07 \pm 0.01$ & $18.03 \pm 0.22$ \\
 & color-blind DINOv3 & $26.17 \pm 0.44$ & $5.09 \pm 0.02$ & $19.98 \pm 0.87$ \\ \midrule

x-harvey-flood
 & DINOv3 & $70.75 \pm 3.70$ & $28.79 \pm 4.82$ & $40.73 \pm 8.93$ \\
 & color-blind DINOv3 & $69.09 \pm 3.86$ & $43.48 \pm 3.21$ & $29.59 \pm 3.52$ \\
\bottomrule
\end{tabular}
\end{table}

\section{Compute Resources}
Our experiments were performed on three machines: DGX A100 and DGX H100 and one HGX H100 node kindly donated by Nebius.ai cloud.

The final \newmodel{} self-supervised pretraining run processed 400 million samples using 8 H100 GPUs for approximately 12 days, consuming roughly 2,300 GPU-hours.

Separately, developing and running the GeoCrossBench benchmark, including hyperparameter searches and repeated or unsuccessful training attempts, consumed 11,900 H100 hours.

\section{All Results} \label{results}
For GeoBench datasets, we report the same metrics as in GeoBench. For other datasets on GeoCrossBench, we adopt the metrics used in previous works. Specifically, for x-sen1floods11, we report mIoU \citep{sen1floods11, pangaea, galileo}. For x-OSCD dataset, we report the F1 score \citep{oscd, gfm}. For x-harvey-building and x-harvey-flood, we first calculated mIoU and bIoU \citep{harvey}. However, because the classes are highly imbalanced, our initial experiments showed that models can achieve a high mIoU simply by perfectly segmenting the majority class while completely failing on the minority class. To avoid this misleading result, we therefore report only the bIoU for the minority class as our evaluation metric.

The following compact tables report the three setting averages for every model on each dataset.

\makeatletter
\setlength{\@fptop}{0pt}
\makeatother
\begin{table*}[h]
\centering
\scriptsize
\renewcommand{\arraystretch}{0.86}
\setlength{\tabcolsep}{2.6pt}
\begin{subtable}[t]{0.485\textwidth}
\centering
\caption{x-bigearthnet}
\label{tab:per-dataset-1}
\begin{tabular}{@{}lccc@{}}
\toprule
Model & ID & No-overlap & Superset \\
\midrule
AnySat & 66.70$\pm$0.36 & 4.64$\pm$1.81 & 2.40$\pm$1.58 \\
AnySat* & 34.08$\pm$0.78 & 10.49$\pm$1.74 & 13.92$\pm$2.58 \\
CROMA & 66.86$\pm$0.14 & 7.53$\pm$0.77 & 62.02$\pm$0.22 \\
CROMA* & 47.89$\pm$3.83 & 12.94$\pm$1.61 & 31.54$\pm$1.59 \\
DINOv2 & 70.27$\pm$0.22 & 3.40$\pm$0.78 & 69.26$\pm$0.29 \\
DINOv2* & 54.61$\pm$0.21 & 16.03$\pm$0.49 & 47.65$\pm$0.56 \\
DINOv3 & 69.75$\pm$0.16 & 7.11$\pm$1.46 & 68.71$\pm$0.41 \\
DINOv3* & 55.09$\pm$0.21 & 15.85$\pm$0.28 & 47.93$\pm$0.31 \\
DOFA & 69.74$\pm$0.23 & 2.25$\pm$0.45 & 53.73$\pm$1.36 \\
DOFA* & 57.49$\pm$0.39 & 4.99$\pm$0.36 & 44.20$\pm$0.43 \\
Panopticon & 69.83$\pm$0.22 & 8.31$\pm$2.30 & 63.19$\pm$0.23 \\
Panopticon* & 62.20$\pm$1.27 & 21.89$\pm$2.48 & 58.15$\pm$0.31 \\
Prithvi & 64.64$\pm$0.19 & 12.16$\pm$0.85 & 63.70$\pm$0.09 \\
Prithvi* & 27.54$\pm$1.69 & 11.12$\pm$1.12 & 24.90$\pm$1.30 \\
ResNet50 & 67.69$\pm$0.38 & 13.62$\pm$2.57 & 66.22$\pm$0.19 \\
ResNet50* & 49.50$\pm$0.37 & 3.88$\pm$0.38 & 38.39$\pm$1.34 \\
SatlasNet & 71.19$\pm$0.38 & 5.71$\pm$2.65 & 69.29$\pm$0.34 \\
SatlasNet* & 15.98$\pm$2.80 & 12.60$\pm$3.09 & 16.76$\pm$1.38 \\
TerraFM & 68.50$\pm$0.22 & 5.07$\pm$1.71 & 55.55$\pm$0.49 \\
TerraFM* & 58.08$\pm$0.35 & 5.96$\pm$0.99 & 39.61$\pm$0.39 \\
TerraMind & 60.49$\pm$1.20 & 9.33$\pm$3.37 & 59.12$\pm$0.93 \\
TerraMind* & 41.98$\pm$0.28 & 10.44$\pm$0.54 & 41.37$\pm$0.34 \\
ViT-B & 69.60$\pm$0.17 & 5.13$\pm$1.50 & 68.48$\pm$0.19 \\
ViT-B* & 51.85$\pm$0.56 & 9.26$\pm$0.81 & 47.36$\pm$0.54 \\
iBOT & 69.54$\pm$0.30 & 4.82$\pm$0.38 & 68.23$\pm$0.25 \\
iBOT* & 57.10$\pm$0.15 & 14.09$\pm$0.27 & 47.76$\pm$0.34 \\
\newmodel{} & 71.76$\pm$0.64 & 10.56$\pm$3.14 & 69.85$\pm$0.71 \\
\newmodel{}* & 67.72$\pm$0.26 & 27.30$\pm$0.40 & 54.59$\pm$0.39 \\
\bottomrule
\end{tabular}
\end{subtable}
\hfill
\begin{subtable}[t]{0.485\textwidth}
\centering
\caption{x-brick-kiln}
\label{tab:per-dataset-2}
\begin{tabular}{@{}lccc@{}}
\toprule
Model & ID & No-overlap & Superset \\
\midrule
AnySat & 98.04$\pm$0.13 & 66.65$\pm$0.06 & 66.37$\pm$0.00 \\
AnySat* & 80.06$\pm$1.04 & 66.88$\pm$2.24 & 63.10$\pm$4.21 \\
CROMA & 98.44$\pm$0.17 & 66.51$\pm$0.13 & 98.25$\pm$0.14 \\
CROMA* & 91.31$\pm$0.34 & 69.21$\pm$0.66 & 76.61$\pm$1.43 \\
DINOv2 & 98.70$\pm$0.12 & 66.63$\pm$0.12 & 98.53$\pm$0.12 \\
DINOv2* & 92.08$\pm$0.29 & 68.56$\pm$0.12 & 90.97$\pm$0.52 \\
DINOv3 & 98.68$\pm$0.24 & 66.81$\pm$0.39 & 98.59$\pm$0.16 \\
DINOv3* & 92.51$\pm$0.14 & 68.58$\pm$0.16 & 91.82$\pm$0.29 \\
DOFA & 98.82$\pm$0.08 & 66.37$\pm$0.00 & 97.84$\pm$0.26 \\
DOFA* & 95.85$\pm$0.17 & 65.99$\pm$0.34 & 88.40$\pm$1.11 \\
Panopticon & 98.67$\pm$0.19 & 67.00$\pm$0.63 & 97.43$\pm$0.42 \\
Panopticon* & 95.36$\pm$0.25 & 73.14$\pm$0.66 & 92.20$\pm$0.19 \\
Prithvi & 97.96$\pm$0.04 & 66.76$\pm$0.38 & 98.07$\pm$0.18 \\
Prithvi* & 86.70$\pm$0.31 & 66.06$\pm$0.11 & 81.78$\pm$0.40 \\
ResNet50 & 98.87$\pm$0.22 & 66.46$\pm$0.23 & 98.71$\pm$0.18 \\
ResNet50* & 87.08$\pm$0.30 & 66.56$\pm$0.81 & 85.91$\pm$0.30 \\
SatlasNet & 98.30$\pm$0.18 & 66.92$\pm$0.27 & 98.31$\pm$0.17 \\
SatlasNet* & 85.31$\pm$0.28 & 68.58$\pm$0.71 & 84.17$\pm$0.30 \\
TerraFM & 98.36$\pm$0.12 & 66.75$\pm$0.46 & 97.08$\pm$0.49 \\
TerraFM* & 92.70$\pm$0.16 & 70.13$\pm$0.46 & 71.31$\pm$1.42 \\
TerraMind & 98.43$\pm$0.20 & 66.46$\pm$0.05 & 98.36$\pm$0.22 \\
TerraMind* & 79.83$\pm$0.77 & 69.11$\pm$0.47 & 85.29$\pm$3.94 \\
ViT-B & 98.66$\pm$0.24 & 66.44$\pm$0.06 & 98.50$\pm$0.14 \\
ViT-B* & 93.06$\pm$0.08 & 67.56$\pm$0.18 & 92.05$\pm$0.15 \\
iBOT & 98.66$\pm$0.17 & 66.70$\pm$0.15 & 98.47$\pm$0.18 \\
iBOT* & 93.74$\pm$0.24 & 69.10$\pm$0.53 & 93.25$\pm$0.36 \\
\newmodel{} & 98.69$\pm$0.08 & 70.40$\pm$1.26 & 98.18$\pm$0.32 \\
\newmodel{}* & 95.45$\pm$0.21 & 67.48$\pm$0.68 & 89.37$\pm$0.62 \\
\bottomrule
\end{tabular}
\end{subtable}
\caption{Per-dataset setting averages for all evaluated models (datasets 1--2 of 10). Values are mean $\pm$ standard deviation across the five seed-level aggregates. An asterisk marks a frozen backbone.}
\end{table*}

\begin{table*}[p]
\centering
\scriptsize
\renewcommand{\arraystretch}{0.86}
\setlength{\tabcolsep}{2.6pt}
\begin{subtable}[t]{0.485\textwidth}
\centering
\caption{x-cashew-plantation}
\label{tab:per-dataset-3}
\begin{tabular}{@{}lccc@{}}
\toprule
Model & ID & No-overlap & Superset \\
\midrule
AnySat & 13.71$\pm$2.44 & 7.01$\pm$0.88 & 13.37$\pm$2.70 \\
AnySat* & 11.75$\pm$0.90 & 5.33$\pm$0.47 & 10.42$\pm$1.56 \\
CROMA & 23.65$\pm$5.17 & 7.23$\pm$0.38 & 20.42$\pm$5.18 \\
CROMA* & 16.97$\pm$0.65 & 6.47$\pm$0.28 & 9.75$\pm$0.49 \\
DINOv2 & 57.61$\pm$2.08 & 9.86$\pm$0.30 & 35.58$\pm$1.18 \\
DINOv2* & 53.03$\pm$0.71 & 8.92$\pm$0.22 & 32.11$\pm$0.94 \\
DINOv3 & 44.03$\pm$3.41 & 8.67$\pm$0.78 & 33.65$\pm$1.53 \\
DINOv3* & 20.69$\pm$0.72 & 6.65$\pm$0.22 & 16.78$\pm$0.58 \\
DOFA & 55.75$\pm$5.05 & 6.06$\pm$0.34 & 35.18$\pm$5.31 \\
DOFA* & 45.32$\pm$0.72 & 6.48$\pm$0.25 & 22.33$\pm$0.76 \\
Panopticon & 44.47$\pm$7.79 & 9.41$\pm$0.97 & 27.53$\pm$4.60 \\
Panopticon* & 45.00$\pm$1.49 & 7.10$\pm$0.97 & 28.44$\pm$1.11 \\
Prithvi & 54.00$\pm$3.24 & 10.03$\pm$1.01 & 47.99$\pm$2.80 \\
Prithvi* & 13.42$\pm$0.73 & 8.11$\pm$0.22 & 11.02$\pm$0.45 \\
ResNet50 & 55.46$\pm$1.29 & 5.97$\pm$0.57 & 38.54$\pm$1.14 \\
ResNet50* & 28.70$\pm$2.09 & 6.47$\pm$0.65 & 16.62$\pm$1.69 \\
SatlasNet & 36.75$\pm$1.85 & 5.75$\pm$0.72 & 26.89$\pm$1.46 \\
SatlasNet* & 18.64$\pm$0.27 & 6.38$\pm$0.19 & 16.41$\pm$0.28 \\
TerraFM & 39.06$\pm$5.15 & 8.23$\pm$0.43 & 26.25$\pm$3.40 \\
TerraFM* & 42.65$\pm$3.86 & 7.31$\pm$1.25 & 14.24$\pm$0.57 \\
TerraMind & 58.65$\pm$4.47 & 6.82$\pm$0.89 & 21.64$\pm$3.58 \\
TerraMind* & 41.23$\pm$1.97 & 5.91$\pm$0.35 & 16.00$\pm$1.27 \\
ViT-B & 52.60$\pm$3.06 & 9.78$\pm$0.86 & 42.29$\pm$2.01 \\
ViT-B* & 43.74$\pm$0.46 & 6.96$\pm$0.38 & 24.56$\pm$0.50 \\
iBOT & 45.14$\pm$3.22 & 7.84$\pm$1.26 & 34.56$\pm$2.20 \\
iBOT* & 30.51$\pm$1.53 & 6.50$\pm$0.27 & 21.78$\pm$1.09 \\
\newmodel{} & 48.11$\pm$3.94 & 11.86$\pm$0.62 & 39.18$\pm$3.06 \\
\newmodel{}* & 44.91$\pm$2.53 & 7.53$\pm$0.93 & 27.59$\pm$1.57 \\
\bottomrule
\end{tabular}
\end{subtable}
\hfill
\begin{subtable}[t]{0.485\textwidth}
\centering
\caption{x-eurosat}
\label{tab:per-dataset-4}
\begin{tabular}{@{}lccc@{}}
\toprule
Model & ID & No-overlap & Superset \\
\midrule
AnySat & 91.71$\pm$0.45 & 12.27$\pm$0.61 & 9.96$\pm$0.12 \\
AnySat* & 71.66$\pm$0.86 & 10.26$\pm$1.10 & 10.72$\pm$1.08 \\
CROMA & 93.59$\pm$0.40 & 18.81$\pm$1.63 & 76.76$\pm$1.35 \\
CROMA* & 86.22$\pm$0.17 & 19.85$\pm$0.13 & 39.89$\pm$0.39 \\
DINOv2 & 97.70$\pm$0.13 & 21.79$\pm$3.30 & 97.33$\pm$0.30 \\
DINOv2* & 90.10$\pm$0.48 & 29.41$\pm$0.49 & 80.15$\pm$1.06 \\
DINOv3 & 97.22$\pm$0.22 & 23.30$\pm$1.30 & 95.47$\pm$0.92 \\
DINOv3* & 92.44$\pm$0.16 & 32.52$\pm$0.43 & 84.31$\pm$0.54 \\
DOFA & 97.24$\pm$0.32 & 12.05$\pm$1.17 & 92.17$\pm$2.22 \\
DOFA* & 94.29$\pm$0.31 & 15.49$\pm$1.15 & 84.71$\pm$0.77 \\
Panopticon & 97.46$\pm$0.30 & 27.87$\pm$1.92 & 96.40$\pm$0.27 \\
Panopticon* & 96.05$\pm$0.22 & 37.83$\pm$0.95 & 94.63$\pm$0.14 \\
Prithvi & 90.78$\pm$0.53 & 13.58$\pm$0.76 & 87.95$\pm$0.40 \\
Prithvi* & 64.44$\pm$0.19 & 14.94$\pm$0.35 & 60.10$\pm$0.53 \\
ResNet50 & 97.44$\pm$0.18 & 18.61$\pm$2.42 & 95.93$\pm$0.20 \\
ResNet50* & 89.09$\pm$0.38 & 15.10$\pm$0.38 & 72.79$\pm$0.97 \\
SatlasNet & 97.11$\pm$0.20 & 15.52$\pm$1.64 & 95.75$\pm$1.49 \\
SatlasNet* & 73.47$\pm$0.39 & 12.56$\pm$0.56 & 72.25$\pm$0.23 \\
TerraFM & 96.88$\pm$0.10 & 15.72$\pm$1.40 & 83.08$\pm$1.73 \\
TerraFM* & 93.39$\pm$0.27 & 16.91$\pm$0.56 & 62.43$\pm$1.12 \\
TerraMind & 91.15$\pm$2.09 & 18.75$\pm$2.48 & 86.08$\pm$2.46 \\
TerraMind* & 73.58$\pm$1.06 & 18.56$\pm$0.27 & 70.72$\pm$1.70 \\
ViT-B & 97.37$\pm$0.14 & 27.33$\pm$1.19 & 96.44$\pm$0.34 \\
ViT-B* & 89.00$\pm$0.36 & 26.67$\pm$0.66 & 79.19$\pm$0.35 \\
iBOT & 97.48$\pm$0.27 & 20.63$\pm$2.94 & 96.38$\pm$0.52 \\
iBOT* & 93.10$\pm$0.33 & 20.11$\pm$1.35 & 77.30$\pm$0.36 \\
\newmodel{} & 98.25$\pm$0.23 & 18.86$\pm$3.39 & 97.79$\pm$0.34 \\
\newmodel{}* & 95.57$\pm$0.26 & 19.89$\pm$0.46 & 92.36$\pm$0.42 \\
\bottomrule
\end{tabular}
\end{subtable}
\caption{Per-dataset setting averages for all evaluated models (datasets 3--4 of 10). Values are mean $\pm$ standard deviation across the five seed-level aggregates. An asterisk marks a frozen backbone.}
\end{table*}

\begin{table*}[p]
\centering
\scriptsize
\renewcommand{\arraystretch}{0.86}
\setlength{\tabcolsep}{2.6pt}
\begin{subtable}[t]{0.485\textwidth}
\centering
\caption{x-harvey-building}
\label{tab:per-dataset-5}
\begin{tabular}{@{}lccc@{}}
\toprule
Model & ID & No-overlap & Superset \\
\midrule
AnySat & 32.49$\pm$9.42 & 7.61$\pm$4.69 & 17.02$\pm$10.33 \\
AnySat* & 32.60$\pm$3.47 & 10.06$\pm$1.50 & 19.16$\pm$1.20 \\
CROMA & 45.55$\pm$5.47 & 17.80$\pm$1.30 & 38.63$\pm$4.55 \\
CROMA* & 39.16$\pm$2.15 & 12.16$\pm$2.57 & 26.98$\pm$4.49 \\
DINOv2 & 43.22$\pm$1.31 & 24.07$\pm$1.55 & 31.50$\pm$2.56 \\
DINOv2* & 41.73$\pm$0.54 & 26.34$\pm$1.83 & 35.20$\pm$1.51 \\
DINOv3 & 41.28$\pm$2.40 & 18.95$\pm$1.25 & 40.79$\pm$2.57 \\
DINOv3* & 32.53$\pm$2.43 & 21.08$\pm$5.08 & 26.04$\pm$4.51 \\
DOFA & 51.93$\pm$1.46 & 19.74$\pm$2.03 & 43.51$\pm$3.54 \\
DOFA* & 41.77$\pm$0.76 & 18.71$\pm$3.48 & 34.38$\pm$3.54 \\
Panopticon & 49.79$\pm$2.08 & 26.01$\pm$3.95 & 43.54$\pm$2.50 \\
Panopticon* & 44.75$\pm$1.08 & 27.99$\pm$3.57 & 36.21$\pm$3.18 \\
Prithvi & 42.68$\pm$0.99 & 21.35$\pm$0.59 & 41.69$\pm$0.96 \\
Prithvi* & 32.47$\pm$2.14 & 25.72$\pm$8.30 & 29.14$\pm$4.60 \\
ResNet50 & 56.92$\pm$0.83 & 20.48$\pm$5.82 & 51.16$\pm$1.75 \\
ResNet50* & 49.06$\pm$0.59 & 23.88$\pm$5.80 & 45.51$\pm$0.57 \\
SatlasNet & 59.27$\pm$0.45 & 11.21$\pm$2.42 & 42.00$\pm$1.34 \\
SatlasNet* & 46.96$\pm$1.32 & 11.45$\pm$0.98 & 24.92$\pm$2.10 \\
TerraFM & 49.48$\pm$0.70 & 16.05$\pm$1.39 & 39.70$\pm$1.79 \\
TerraFM* & 44.91$\pm$0.61 & 15.72$\pm$4.04 & 28.78$\pm$2.55 \\
TerraMind & 48.86$\pm$0.91 & 19.78$\pm$5.02 & 32.73$\pm$7.82 \\
TerraMind* & 44.11$\pm$0.79 & 11.13$\pm$5.30 & 27.03$\pm$7.91 \\
ViT-B & 48.93$\pm$1.51 & 25.36$\pm$2.79 & 48.52$\pm$1.42 \\
ViT-B* & 39.70$\pm$0.91 & 19.33$\pm$2.94 & 37.35$\pm$1.45 \\
iBOT & 46.31$\pm$1.95 & 20.52$\pm$1.78 & 46.26$\pm$1.95 \\
iBOT* & 39.81$\pm$0.81 & 12.76$\pm$1.63 & 36.14$\pm$0.95 \\
\newmodel{} & 49.03$\pm$0.68 & 24.34$\pm$2.39 & 46.41$\pm$0.72 \\
\newmodel{}* & 39.84$\pm$1.73 & 26.33$\pm$3.45 & 30.17$\pm$5.51 \\
\bottomrule
\end{tabular}
\end{subtable}
\hfill
\begin{subtable}[t]{0.485\textwidth}
\centering
\caption{x-harvey-flood}
\label{tab:per-dataset-6}
\begin{tabular}{@{}lccc@{}}
\toprule
Model & ID & No-overlap & Superset \\
\midrule
AnySat & 61.35$\pm$3.47 & 17.35$\pm$1.21 & 11.53$\pm$1.84 \\
AnySat* & 50.92$\pm$15.25 & 6.81$\pm$1.12 & 7.27$\pm$3.15 \\
CROMA & 45.37$\pm$13.70 & 7.81$\pm$6.80 & 20.14$\pm$9.38 \\
CROMA* & 45.14$\pm$0.76 & 7.33$\pm$0.54 & 17.10$\pm$1.51 \\
DINOv2 & 82.73$\pm$0.11 & 25.96$\pm$0.12 & 45.06$\pm$0.10 \\
DINOv2* & 77.10$\pm$0.35 & 20.67$\pm$0.76 & 38.34$\pm$0.81 \\
DINOv3 & 67.30$\pm$0.35 & 17.96$\pm$0.89 & 59.76$\pm$0.48 \\
DINOv3* & 46.34$\pm$0.82 & 3.20$\pm$0.89 & 32.48$\pm$2.08 \\
DOFA & 74.28$\pm$0.45 & 0.27$\pm$0.26 & 56.66$\pm$0.84 \\
DOFA* & 57.48$\pm$0.19 & 2.55$\pm$0.65 & 31.82$\pm$0.39 \\
Panopticon & 85.17$\pm$4.88 & 19.82$\pm$1.77 & 63.88$\pm$6.48 \\
Panopticon* & 77.59$\pm$0.30 & 5.98$\pm$0.87 & 33.81$\pm$0.44 \\
Prithvi & 65.36$\pm$2.12 & 16.76$\pm$3.91 & 64.15$\pm$2.03 \\
Prithvi* & 49.31$\pm$1.37 & 6.31$\pm$1.23 & 39.92$\pm$1.19 \\
ResNet50 & 61.49$\pm$1.08 & 13.60$\pm$1.93 & 39.28$\pm$2.81 \\
ResNet50* & 65.69$\pm$0.49 & 9.39$\pm$1.72 & 29.77$\pm$0.81 \\
SatlasNet & 57.40$\pm$1.11 & 10.21$\pm$1.27 & 35.15$\pm$1.80 \\
SatlasNet* & 41.37$\pm$0.29 & 5.78$\pm$1.02 & 27.81$\pm$1.04 \\
TerraFM & 70.07$\pm$0.18 & 23.63$\pm$0.80 & 33.61$\pm$2.16 \\
TerraFM* & 62.82$\pm$0.29 & 13.12$\pm$1.59 & 24.76$\pm$2.01 \\
TerraMind & 70.88$\pm$1.26 & 10.74$\pm$0.53 & 25.93$\pm$0.54 \\
TerraMind* & 63.59$\pm$0.31 & 6.95$\pm$1.37 & 16.25$\pm$1.16 \\
ViT-B & 73.48$\pm$1.49 & 37.58$\pm$2.19 & 67.96$\pm$1.35 \\
ViT-B* & 56.43$\pm$0.28 & 15.18$\pm$1.04 & 30.94$\pm$1.87 \\
iBOT & 68.84$\pm$0.37 & 19.63$\pm$0.63 & 59.32$\pm$0.55 \\
iBOT* & 55.46$\pm$0.56 & 5.00$\pm$1.13 & 34.88$\pm$0.79 \\
\newmodel{} & 55.83$\pm$8.41 & 22.71$\pm$3.82 & 47.57$\pm$8.55 \\
\newmodel{}* & 42.80$\pm$0.88 & 5.47$\pm$1.87 & 25.61$\pm$3.62 \\
\bottomrule
\end{tabular}
\end{subtable}
\caption{Per-dataset setting averages for all evaluated models (datasets 5--6 of 10). Values are mean $\pm$ standard deviation across the five seed-level aggregates. An asterisk marks a frozen backbone.}
\end{table*}

\begin{table*}[p]
\centering
\scriptsize
\renewcommand{\arraystretch}{0.86}
\setlength{\tabcolsep}{2.6pt}
\begin{subtable}[t]{0.485\textwidth}
\centering
\caption{x-OSCD}
\label{tab:per-dataset-7}
\begin{tabular}{@{}lccc@{}}
\toprule
Model & ID & No-overlap & Superset \\
\midrule
AnySat & 9.70$\pm$7.59 & 6.82$\pm$2.44 & 4.67$\pm$1.71 \\
AnySat* & 7.10$\pm$1.47 & 7.10$\pm$0.70 & 5.07$\pm$2.11 \\
CROMA & 10.40$\pm$1.96 & 8.19$\pm$3.26 & 12.67$\pm$7.36 \\
CROMA* & 18.82$\pm$5.34 & 12.44$\pm$2.74 & 15.88$\pm$1.73 \\
DINOv2 & 34.43$\pm$4.24 & 9.25$\pm$2.84 & 33.31$\pm$2.79 \\
DINOv2* & 28.42$\pm$2.92 & 8.84$\pm$2.94 & 30.29$\pm$2.69 \\
DINOv3 & 7.62$\pm$1.09 & 7.19$\pm$1.22 & 6.93$\pm$1.22 \\
DINOv3* & 8.21$\pm$1.69 & 7.36$\pm$1.17 & 7.26$\pm$2.32 \\
DOFA & 23.13$\pm$5.19 & 5.78$\pm$2.12 & 17.98$\pm$3.52 \\
DOFA* & 24.41$\pm$7.04 & 3.78$\pm$1.77 & 17.78$\pm$7.34 \\
Panopticon & 21.34$\pm$5.91 & 13.31$\pm$0.86 & 19.07$\pm$6.32 \\
Panopticon* & 32.27$\pm$2.48 & 14.71$\pm$1.01 & 25.63$\pm$2.45 \\
Prithvi & 13.64$\pm$4.99 & 5.87$\pm$2.96 & 7.09$\pm$2.01 \\
Prithvi* & 1.62$\pm$1.30 & 1.93$\pm$1.27 & 2.72$\pm$1.13 \\
ResNet50 & 20.93$\pm$4.45 & 9.52$\pm$1.94 & 26.99$\pm$4.01 \\
ResNet50* & 19.96$\pm$4.65 & 9.20$\pm$0.48 & 26.60$\pm$4.13 \\
SatlasNet & 23.91$\pm$7.98 & 11.08$\pm$2.17 & 21.15$\pm$6.69 \\
SatlasNet* & 16.46$\pm$2.21 & 9.08$\pm$0.82 & 16.66$\pm$1.31 \\
TerraFM & 34.81$\pm$2.25 & 13.61$\pm$1.66 & 29.84$\pm$4.56 \\
TerraFM* & 31.10$\pm$3.17 & 11.71$\pm$1.18 & 28.56$\pm$5.73 \\
TerraMind & 32.70$\pm$3.67 & 9.21$\pm$1.69 & 14.41$\pm$3.57 \\
TerraMind* & 31.32$\pm$2.55 & 13.04$\pm$3.66 & 13.82$\pm$0.52 \\
ViT-B & 20.16$\pm$7.89 & 8.33$\pm$1.93 & 16.65$\pm$8.27 \\
ViT-B* & 25.35$\pm$3.76 & 9.01$\pm$2.53 & 27.57$\pm$3.84 \\
iBOT & 11.87$\pm$7.58 & 7.29$\pm$2.09 & 8.27$\pm$5.69 \\
iBOT* & 8.07$\pm$4.11 & 6.73$\pm$0.48 & 8.43$\pm$5.17 \\
\newmodel{} & 15.79$\pm$2.68 & 15.08$\pm$3.38 & 17.23$\pm$4.42 \\
\newmodel{}* & 12.88$\pm$2.68 & 11.92$\pm$2.32 & 6.84$\pm$2.39 \\
\bottomrule
\end{tabular}
\end{subtable}
\hfill
\begin{subtable}[t]{0.485\textwidth}
\centering
\caption{x-SA-crop-type}
\label{tab:per-dataset-8}
\begin{tabular}{@{}lccc@{}}
\toprule
Model & ID & No-overlap & Superset \\
\midrule
AnySat & 8.75$\pm$0.85 & 5.48$\pm$0.44 & 10.40$\pm$1.34 \\
AnySat* & 8.10$\pm$0.18 & 5.13$\pm$0.74 & 8.23$\pm$0.45 \\
CROMA & 17.98$\pm$1.58 & 4.89$\pm$0.42 & 14.83$\pm$1.11 \\
CROMA* & 15.83$\pm$1.69 & 4.01$\pm$0.61 & 11.08$\pm$0.89 \\
DINOv2 & 18.96$\pm$0.80 & 7.61$\pm$0.45 & 16.96$\pm$0.83 \\
DINOv2* & 20.05$\pm$0.47 & 7.74$\pm$0.47 & 18.33$\pm$0.50 \\
DINOv3 & 10.89$\pm$0.75 & 5.23$\pm$0.10 & 10.63$\pm$0.74 \\
DINOv3* & 10.23$\pm$0.40 & 4.40$\pm$0.30 & 9.85$\pm$0.51 \\
DOFA & 23.26$\pm$0.83 & 5.51$\pm$0.36 & 15.09$\pm$0.97 \\
DOFA* & 21.45$\pm$0.49 & 6.39$\pm$0.45 & 17.25$\pm$0.29 \\
Panopticon & 15.86$\pm$1.50 & 6.72$\pm$0.30 & 15.12$\pm$1.27 \\
Panopticon* & 21.52$\pm$1.04 & 7.06$\pm$0.43 & 19.57$\pm$0.95 \\
Prithvi & 20.16$\pm$1.10 & 6.37$\pm$0.16 & 20.16$\pm$1.10 \\
Prithvi* & 9.13$\pm$0.81 & 4.91$\pm$0.48 & 9.15$\pm$0.95 \\
ResNet50 & 20.38$\pm$1.23 & 7.69$\pm$0.46 & 20.12$\pm$1.20 \\
ResNet50* & 16.67$\pm$0.80 & 6.85$\pm$0.28 & 14.24$\pm$0.85 \\
SatlasNet & 15.08$\pm$2.00 & 5.44$\pm$0.53 & 14.96$\pm$2.05 \\
SatlasNet* & 10.44$\pm$1.22 & 5.02$\pm$0.30 & 10.01$\pm$1.11 \\
TerraFM & 18.89$\pm$1.33 & 5.92$\pm$0.71 & 15.82$\pm$1.50 \\
TerraFM* & 20.84$\pm$0.67 & 7.44$\pm$0.15 & 15.17$\pm$0.68 \\
TerraMind & 23.63$\pm$0.88 & 5.38$\pm$0.44 & 11.41$\pm$1.20 \\
TerraMind* & 18.98$\pm$0.74 & 5.35$\pm$0.17 & 10.09$\pm$0.17 \\
ViT-B & 20.78$\pm$2.70 & 6.62$\pm$0.35 & 20.37$\pm$2.54 \\
ViT-B* & 19.52$\pm$0.49 & 6.93$\pm$0.25 & 17.35$\pm$0.56 \\
iBOT & 14.62$\pm$0.53 & 5.39$\pm$0.54 & 14.31$\pm$0.67 \\
iBOT* & 10.98$\pm$0.67 & 4.46$\pm$0.64 & 10.66$\pm$0.86 \\
\newmodel{} & 15.91$\pm$2.04 & 7.36$\pm$0.12 & 15.29$\pm$1.94 \\
\newmodel{}* & 18.45$\pm$1.04 & 7.76$\pm$0.52 & 16.21$\pm$0.90 \\
\bottomrule
\end{tabular}
\end{subtable}
\caption{Per-dataset setting averages for all evaluated models (datasets 7--8 of 10). Values are mean $\pm$ standard deviation across the five seed-level aggregates. An asterisk marks a frozen backbone.}
\end{table*}

\begin{table*}[p]
\centering
\scriptsize
\renewcommand{\arraystretch}{0.86}
\setlength{\tabcolsep}{2.6pt}
\begin{subtable}[t]{0.485\textwidth}
\centering
\caption{x-sen1floods11}
\label{tab:per-dataset-9}
\begin{tabular}{@{}lccc@{}}
\toprule
Model & ID & No-overlap & Superset \\
\midrule
AnySat & 7.97$\pm$4.10 & 4.95$\pm$3.33 & 16.28$\pm$8.73 \\
AnySat* & 10.55$\pm$2.05 & 6.35$\pm$3.74 & 14.32$\pm$7.07 \\
CROMA & 46.31$\pm$10.07 & 5.70$\pm$1.51 & 36.36$\pm$8.00 \\
CROMA* & 42.36$\pm$3.28 & 4.48$\pm$1.36 & 24.81$\pm$2.69 \\
DINOv2 & 47.18$\pm$6.26 & 15.38$\pm$4.50 & 34.41$\pm$6.84 \\
DINOv2* & 52.14$\pm$1.74 & 17.84$\pm$3.99 & 39.24$\pm$3.78 \\
DINOv3 & 40.57$\pm$4.66 & 12.70$\pm$2.45 & 40.76$\pm$5.77 \\
DINOv3* & 25.56$\pm$2.24 & 7.42$\pm$4.30 & 29.12$\pm$4.26 \\
DOFA & 59.02$\pm$2.48 & 7.64$\pm$5.67 & 45.98$\pm$4.58 \\
DOFA* & 58.27$\pm$0.27 & 9.49$\pm$2.75 & 44.11$\pm$5.54 \\
Panopticon & 57.95$\pm$2.50 & 19.91$\pm$5.80 & 58.94$\pm$4.26 \\
Panopticon* & 57.23$\pm$0.75 & 15.47$\pm$4.97 & 59.20$\pm$4.13 \\
Prithvi & 36.61$\pm$2.45 & 5.73$\pm$3.72 & 36.27$\pm$2.41 \\
Prithvi* & 23.57$\pm$4.02 & 4.10$\pm$2.88 & 26.24$\pm$5.15 \\
ResNet50 & 60.61$\pm$3.23 & 13.26$\pm$8.57 & 55.35$\pm$3.53 \\
ResNet50* & 51.17$\pm$1.32 & 20.47$\pm$5.10 & 47.28$\pm$2.45 \\
SatlasNet & 51.67$\pm$3.59 & 6.01$\pm$3.29 & 47.14$\pm$0.73 \\
SatlasNet* & 46.92$\pm$2.63 & 6.15$\pm$3.48 & 44.26$\pm$2.91 \\
TerraFM & 54.73$\pm$5.68 & 13.01$\pm$3.66 & 25.57$\pm$2.46 \\
TerraFM* & 56.23$\pm$1.02 & 12.94$\pm$3.51 & 34.98$\pm$2.50 \\
TerraMind & 51.86$\pm$8.20 & 6.68$\pm$4.10 & 19.17$\pm$1.90 \\
TerraMind* & 50.77$\pm$6.02 & 5.71$\pm$3.31 & 13.23$\pm$4.21 \\
ViT-B & 50.51$\pm$11.16 & 11.27$\pm$4.04 & 48.94$\pm$9.86 \\
ViT-B* & 51.12$\pm$1.88 & 5.58$\pm$1.68 & 32.74$\pm$2.97 \\
iBOT & 39.01$\pm$8.64 & 8.51$\pm$5.43 & 34.94$\pm$7.77 \\
iBOT* & 31.69$\pm$3.97 & 3.67$\pm$2.95 & 33.38$\pm$1.82 \\
\newmodel{} & 60.56$\pm$2.83 & 34.46$\pm$4.18 & 60.61$\pm$1.99 \\
\newmodel{}* & 57.61$\pm$1.47 & 33.50$\pm$5.37 & 61.32$\pm$1.73 \\
\bottomrule
\end{tabular}
\end{subtable}
\hfill
\begin{subtable}[t]{0.485\textwidth}
\centering
\caption{x-so2sat}
\label{tab:per-dataset-10}
\begin{tabular}{@{}lccc@{}}
\toprule
Model & ID & No-overlap & Superset \\
\midrule
AnySat & 45.39$\pm$1.49 & 6.42$\pm$0.80 & 6.82$\pm$0.43 \\
AnySat* & 30.81$\pm$0.90 & 6.26$\pm$1.04 & 4.56$\pm$0.87 \\
CROMA & 50.46$\pm$0.78 & 7.27$\pm$0.48 & 32.97$\pm$3.58 \\
CROMA* & 45.86$\pm$1.26 & 7.28$\pm$0.22 & 15.04$\pm$0.67 \\
DINOv2 & 59.09$\pm$1.01 & 8.90$\pm$0.68 & 56.12$\pm$0.94 \\
DINOv2* & 51.21$\pm$0.88 & 12.26$\pm$0.28 & 39.45$\pm$0.57 \\
DINOv3 & 59.25$\pm$1.22 & 8.82$\pm$0.68 & 54.30$\pm$1.23 \\
DINOv3* & 52.32$\pm$0.42 & 10.89$\pm$0.21 & 38.50$\pm$0.53 \\
DOFA & 59.83$\pm$0.74 & 5.90$\pm$0.76 & 43.36$\pm$6.29 \\
DOFA* & 58.06$\pm$0.78 & 7.04$\pm$0.46 & 36.27$\pm$1.35 \\
Panopticon & 59.67$\pm$1.75 & 12.45$\pm$1.49 & 57.56$\pm$1.32 \\
Panopticon* & 62.27$\pm$1.05 & 16.29$\pm$0.80 & 61.38$\pm$0.56 \\
Prithvi & 47.05$\pm$0.41 & 6.51$\pm$0.86 & 42.01$\pm$1.05 \\
Prithvi* & 30.29$\pm$0.55 & 6.99$\pm$0.56 & 23.18$\pm$0.79 \\
ResNet50 & 55.15$\pm$1.10 & 8.26$\pm$0.35 & 50.15$\pm$0.69 \\
ResNet50* & 43.57$\pm$0.49 & 8.33$\pm$0.26 & 26.56$\pm$0.93 \\
SatlasNet & 58.12$\pm$1.08 & 6.54$\pm$0.52 & 53.02$\pm$2.04 \\
SatlasNet* & 35.00$\pm$0.52 & 5.61$\pm$0.45 & 30.23$\pm$0.84 \\
TerraFM & 54.70$\pm$0.91 & 7.71$\pm$0.66 & 34.83$\pm$4.89 \\
TerraFM* & 51.37$\pm$0.68 & 6.42$\pm$0.18 & 22.11$\pm$0.24 \\
TerraMind & 47.05$\pm$3.11 & 7.69$\pm$1.42 & 42.48$\pm$4.17 \\
TerraMind* & 39.92$\pm$0.52 & 6.84$\pm$0.39 & 34.25$\pm$0.58 \\
ViT-B & 57.90$\pm$1.70 & 8.23$\pm$0.45 & 53.76$\pm$1.64 \\
ViT-B* & 51.03$\pm$0.48 & 9.32$\pm$0.70 & 34.03$\pm$0.69 \\
iBOT & 58.76$\pm$1.20 & 8.23$\pm$0.74 & 53.80$\pm$0.31 \\
iBOT* & 52.63$\pm$0.32 & 7.24$\pm$0.10 & 31.98$\pm$0.38 \\
\newmodel{} & 60.97$\pm$1.79 & 10.24$\pm$1.54 & 56.21$\pm$2.78 \\
\newmodel{}* & 58.90$\pm$0.42 & 11.43$\pm$0.45 & 47.60$\pm$0.80 \\
\bottomrule
\end{tabular}
\end{subtable}
\caption{Per-dataset setting averages for all evaluated models (datasets 9--10 of 10). Values are mean $\pm$ standard deviation across the five seed-level aggregates. An asterisk marks a frozen backbone.}
\end{table*}

\end{document}